\documentclass[sigconf]{acmart}

\usepackage{booktabs} 
\usepackage{amsmath,amssymb}
\usepackage{algorithm}
\usepackage{algorithmic}
\usepackage{graphicx} 
\usepackage{subfigure} 
\usepackage{verbatim}
\usepackage{psfrag}
\usepackage{xspace}

\newcommand{\diff}[2]{\frac{\partial #1}{\partial #2}}
\newcommand{\vct}[1]{\ensuremath{\boldsymbol{#1}}} 

\newcommand{\set}[1]{\ensuremath{\mathcal{#1}}}

\newcommand{\T}{\ensuremath{\top}}

\newcommand{\argmax}{\operatornamewithlimits{\arg\,\max}}

\newcommand{\argmin}{\operatornamewithlimits{\arg\,\min}}

\newcommand{\myparagraph}[1]{\smallskip \noindent \textbf{#1.}}

\newcommand{\ie}{{i.e.}\xspace}
\newcommand{\eg}{{e.g.}\xspace}

\newcommand{\aka}{{a.k.a.}\xspace}

\setcopyright{none} 

\begin{document}
\title{Towards Poisoning of Deep Learning Algorithms with Back-gradient Optimization}

\author{Luis Mu\~noz-Gonz\'alez}
\affiliation{Imperial College London, UK}

\author{Battista Biggio}
\affiliation{DIEE, University of Cagliari, Italy}
\affiliation{Pluribus One}

\author{Ambra Demontis}
\affiliation{DIEE, University of Cagliari, Italy}

\author{Andrea Paudice}\affiliation{Imperial College London, UK}
\author{Vasin Wongrassamee}\affiliation{Imperial College London, UK}
\author{Emil C. Lupu} \affiliation{Imperial College London, UK}

\author{Fabio Roli}
\affiliation{DIEE, University of Cagliari, Italy}
\affiliation{Pluribus One}

\begin{abstract}
A number of online services nowadays rely upon machine learning to extract valuable information from data  collected in the wild. 
This exposes learning algorithms to the threat of data poisoning, i.e., a coordinate attack in which a fraction of the training data is controlled by the attacker and manipulated to subvert the learning process.
To date, these attacks have been devised only against a limited class of binary learning algorithms, due to the inherent complexity of the gradient-based procedure used to optimize the poisoning points (\aka adversarial training examples).
In this work, we first extend the definition of poisoning attacks to multiclass problems.
We then propose a novel poisoning algorithm based on the idea of back-gradient optimization, \ie, to compute the gradient of interest through automatic differentiation, while also reversing the learning procedure to drastically reduce the attack complexity.
Compared to current poisoning strategies, our approach is able to target a wider class of learning algorithms, trained with gradient-based procedures, including neural networks and deep learning architectures.
We empirically evaluate its effectiveness on several application examples, including spam filtering, malware detection, and handwritten digit recognition. 
We finally show that, similarly to adversarial test examples, adversarial training examples can also be transferred across different learning algorithms.
\end{abstract}


\maketitle

\section{Introduction}
\label{sec:Introduction}
In recent years technology has become pervasive, enabling a rapid a disruptive change in the way society is organized. Our data is provided to third-party services which are supposed to facilitate and protect our daily work and activities. Most of these services leverage machine learning to extract valuable information from the overwhelming amount of input data received. Although this provides advantages to the users themselves, \eg, in terms of usability and functionality of such services, it is also clear that these services may be abused, providing great opportunities for cybercriminals to conduct novel, illicit, and highly-profitable activities.
Being one of the main components behind such services makes machine learning an appealing target for attackers, who may gain a significant advantage by gaming the learning algorithm. 
Notably, machine learning itself can be the \emph{weakest link} in the security chain, as its vulnerabilities can be exploited by the attacker to compromise the whole system infrastructure. To this end, she may inject malicious data to poison the learning process, or manipulate data at test time to evade detection.\footnote{We refer to the attacker here as feminine due to the common interpretation as ``Eve'' or ``Carol'' in cryptography and security.}
These kinds of attack have been reported against anti-virus engines, anti-spam filters, and systems aimed to detect fake profiles or news in social networks -- all problems involving a well-crafted deployment of machine learning algorithms~\cite{nelson,rubinstein09,biggio12-icml,srndic14,biggio15-icml,mei15-aaai,kloft12b,huang11,joseph13-dagstuhl,smutz12,wang14-usenix}. Such attacks have fuelled a growing interest in the research area of \emph{adversarial machine learning}, at the intersection of cybersecurity and machine learning. This recent research field aims at understanding the security properties of current learning algorithms, as well as at developing more secure ones~\cite{huang11,joseph13-dagstuhl,biggio14-tkde}.

Among the different attack scenarios envisaged against machine learning, \emph{poisoning attacks} are considered one of the most relevant and emerging security threats for data-driven technologies, \ie, technologies relying upon the collection of large amounts of data in the wild~\cite{joseph13-dagstuhl}. In a poisoning attack, the attacker is assumed to control a fraction of the training data used by the learning algorithm, with the goal of subverting the entire learning process, or facilitate subsequent system evasion~\cite{nelson,rubinstein09,biggio12-icml,biggio15-icml,mei15-aaai,kloft12b}.
More practically, data poisoning is already a relevant threat in different application domains.
For instance, some online services directly exploit users' feedback on their decisions to update the trained model. PDFRate\footnote{\url{http://pdfrate.com}} is an online malware detection tool that analyzes the submitted PDF files to reveal the presence of embedded malware~\cite{smutz12}.
After classification, it allows the user to provide feedback on its decision, \ie, to confirm or not the classification result. A malicious user may thus provide wrong feedback to gradually poison the system and compromise its performance over time.
Notably, there is a more general underlying problem related to the collection of large data volumes with reliable labels. This is a well-known problem in malware detection, where malware samples are collected by means of compromised machines with known vulnerabilities (\ie, honeypots), or via other online services, like VirusTotal,\footnote{\url{https://virustotal.com}} in which labelling errors are often reported.

Previous work has developed poisoning attacks against popular learning algorithms like Support Vector Machines (SVMs), LASSO, logistic and ridge regression, in different applications, like spam and malware detection~\cite{nelson,rubinstein09,kloft12b,biggio12-icml,biggio15-icml,mei15-aaai,koh17-icml}.
The main technical difficulty in devising a poisoning attack is the computation of the poisoning samples,
also recently referred to as \emph{adversarial training examples}~\cite{koh17-icml}. This requires solving a bilevel optimization problem in which the outer optimization amounts to maximizing the classification error on an untainted validation set, while the inner optimization corresponds to training the learning algorithm on the poisoned data~\cite{mei15-aaai}.
Since solving this problem with black-box optimization is too computationally demanding,
previous work has exploited gradient-based optimization, along with the idea of \emph{implicit differentiation}.
The latter consists of replacing the inner optimization problem with its stationarity (Karush-Kuhn-Tucker, KKT) conditions to derive an implicit equation for the gradient~\cite{biggio12-icml,biggio15-icml,mei15-aaai,koh17-icml}. 
This approach however can only be used against a limited class of learning algorithms, excluding neural networks and deep learning architectures, due to the inherent complexity of the procedure used to compute the required gradient.
Another limitation is that, to date, previous work has only considered poisoning of two-class learning algorithms. 

In this work, we overcome these limitations by first extending the threat model proposed in~\cite{barreno06-asiaccs,barreno10,huang11,biggio14-tkde} to account for multiclass poisoning attacks (Sect.~\ref{sec:attackerModel}). 
We then exploit a recent technique called \emph{back-gradient optimization}, originally proposed for hyperparameter optimization~\cite{bengio2000,domke12-aistats,maclaurin15-icml,pedregosa16-icml}, to implement a much more computationally-efficient poisoning attack.
The underlying idea is to compute the gradient of interest through reverse-mode (automatic) differentiation (\ie, \emph{back-propagation}), while reversing the underlying learning procedure to 
trace back the entire sequence of parameter updates performed during learning, without storing it. 
In fact, storing this sequence in memory would be infeasible for learning algorithms that optimize a large set of parameters across several iterations.
Our poisoning algorithm only requires the learning algorithm to update its parameters during training in a \emph{smooth} manner (e.g., through gradient descent), to correctly trace these changes backwards. 
Accordingly, compared to previously proposed poisoning strategies, our approach is the first capable of targeting a wider class of learning algorithms, trainable with gradient-based procedures, like neural networks and deep learning architectures (Sect.~\ref{sec:RMD}). 

Another important contribution of this work is to show how the performance of learning algorithms may be drastically compromised even by the presence of a small fraction of poisoning points in the training data, in the context of real-world applications like spam filtering, malware detection, and handwritten digit recognition (Sect.~\ref{sec:Experiments}). 
We also investigate the \emph{transferability} property of poisoning attacks, \ie, the extent to which attacks devised against a specific learning algorithm are effective against different ones. To our knowledge, this property has been investigated for evasion attacks (\aka adversarial test examples), \ie,
attacks aimed to evade a trained classifier at test time~\cite{biggio13-ecml,srndic14,papernot17-asiaccs,moosavi17-cvpr}, but never for poisoning attacks.
We conclude our work by discussing related work (Sect.~\ref{sec:RelatedWork}), the main limitations of our approach, and future research directions (Sect.~\ref{sec:Conclusion}).

\section{Threat Model}
\label{sec:attackerModel}
In this section, we summarize the framework originally proposed in~\cite{barreno06-asiaccs,barreno10,huang11} and subsequently extended in~\cite{biggio14-tkde}, which enables one to envision different attack scenarios against learning algorithms (including deep learning ones), and to craft the corresponding attack samples.
Remarkably, these include attacks at training and at test time, usually referred to as poisoning and evasion attacks~\cite{huang11,biggio14-tkde,biggio13-ecml,biggio12-icml,biggio15-icml,mei15-aaai} or, more recently, as adversarial (training and test) examples (when crafted against deep learning algorithms)~\cite{szegedy14-iclr,papernot16-sp,papernot17-asiaccs}.

The framework characterizes the attacker according to her goal, knowledge of the targeted system, and capability of manipulating the input data.
Based on these assumptions, it allows one to define an optimal attack strategy as an optimization problem whose solution amounts to the construction of the attack samples, \ie, of the \emph{adversarial examples}.

In this work, we extend this framework, originally developed for binary classification problems, to multiclass classification.
While this generalization holds for evasion attacks too, we only detail here the main poisoning attack scenarios.

\myparagraph{Notation} In a classification task, given the instance space $\set X$ and the label space $\set Y$, the learner aims to estimate the underlying (possibly noisy) latent function $f$ that maps $\set X \mapsto \set Y$. Given a training set $\set D_{\rm tr} = \{\vct x_{i}, y_{i} \}_{i=1}^{n}$ with $n$ i.i.d. samples drawn from the underlying probability distribution $p(\set X, \set Y)$,\footnote{While normally the set notation $\{\vct x_{i}, y_{i} \}_{i=1}^{n}$ does not admit duplicate entries, we admit our data sets to contain potentially duplicated points.}
we can estimate $f$ with a parametric or non-parametric model $\set M$ trained by minimizing an objective function $\set L(\set D, \vct w)$ (normally, a tractable estimate of the generalization error), w.r.t. its parameters and/or hyperparameters $\vct w$.\footnote{For instance, for kernelized SVMs, $\vct w$ may include the dual variables $\vct \alpha$, the bias $b$, and even the regularization parameter $C$. In this work, as in~\cite{biggio12-icml,biggio15-icml,mei15-aaai}, we however consider only the optimization of the model parameters, and not of its hyperparameters.}

Thus, while $\set L$ denotes the learner's objective function (possibly including regularization), we use $L(\set D, \vct w)$ to denote only the \emph{loss} incurred when evaluating the learner parameterized by $\vct w$ on the samples in $\set D$.

\subsection{Attacker's Goal} \label{sect:goal}
The goal of the attack is determined in terms of the desired \textbf{security violation} and \textbf{attack specificity}. In multiclass classification, misclassifying a sample does not have a unique meaning, as there is more than one class different from the correct one. Accordingly, we extend the current framework by introducing the concept of \textbf{error specificity}.
These three characteristics are detailed below.

\myparagraph{Security Violation} This characteristic defines the high-level security violation caused by the attack, as normally done in security engineering. It can be: an \emph{integrity} violation, if malicious activities evade detection without compromising normal system operation; an \emph{availability} violation, if normal system functionality is compromised, \eg, by increasing the classification error; or a \emph{privacy} violation, if the attacker obtains private information about the system, its users or data by reverse-engineering the learning algorithm.

\myparagraph{Attack Specificity} This characteristic ranges from \emph{targeted} to \emph{indiscriminate}, respectively, if the attack aims to cause misclassification of a specific set of samples (to target a given system user or protected service), or of any sample (to target any system user or protected service).

\myparagraph{Error Specificity} We introduce here this characteristic to disambiguate the notion of misclassification in multiclass problems. The error specificity can thus be: \emph{specific}, if the attacker aims to have a sample misclassified as a specific class; or \emph{generic}, if the attacker aims to have a sample misclassified as any of the classes different from the true class.\footnote{In~\cite{papernot16-sp}, the authors defined \emph{targeted} and \emph{indiscriminate} attacks (at test time) depending on whether the attacker aims to cause \emph{specific} or \emph{generic} errors. Here we do not follow their naming convention, as it can cause confusion with the interpretation of \emph{targeted} and \emph{indiscriminate} attacks introduced in previous work~\cite{barreno06-asiaccs,barreno10,huang11,biggio14-tkde,biggio15-icml,biggio13-aisec,biggio14-spr,biggio14-aisec}.}
 
\subsection{Attacker's Knowledge} \label{sect:knowledge}

The attacker can have different levels of knowledge of the targeted system, including: ($k.i$) the training data $\set D_{\rm tr}$; ($k.ii$) the feature set $\set X$; ($k.iii$) the learning algorithm $\set M$, along with the objective function $\set L$ minimized during training; and, possibly, ($k.iv$) its (trained) parameters $\vct w$.
The attacker's knowledge can thus be characterized in terms of a space $\Theta$ that encodes the aforementioned assumptions ($k.i$)-($k.iv$) as $\vct \theta=(\set D, \set X, \set M, \vct w)$.
Depending on the assumptions made on each of these components, one can envisage different attack scenarios. Typically, two main settings are considered, referred to as attacks with \emph{perfect} and \emph{limited} knowledge. 

\myparagraph{Perfect-Knowledge (PK) Attacks} In this case, the attacker is assumed to know everything about the targeted system. Although this setting may be not always representative of practical cases, it enables us to perform a worst-case evaluation of the security of learning algorithms under attack, highlighting the upper bounds on the performance degradation that may be incurred by the system under attack. In this case, we have $\vct \theta_{\rm PK}=(\set D, \set X, \set M, \vct w)$.

\myparagraph{Limited-Knowledge (LK) Attacks} Although LK attacks admit a wide range of possibilities, the attacker is typically assumed to know the feature representation $\set X$ and the learning algorithm $\set M$, but not the training data (for which surrogate data from similar sources can be collected). We refer to this case here as {LK attacks with Surrogate Data} (LK-SD), and denote it with $\vct \theta_{\rm LK-SD}=(\hat{\set D}, \set X, \set M, \hat{\vct w})$ (using the \emph{hat} symbol to denote limited knowledge of a given component).
Notably, in this case, as the attacker is only given a surrogate data set $\hat{\set D}$, also the learner's parameters have to be estimated by the attacker, \eg, by optimizing $\set L$ on $\hat{\set D}$.

Similarly, we refer to the case in which the attacker knows the training data (\eg, if the learning algorithm is trained on publicly-available data), but not the learning algorithm (for which a surrogate learner can be trained on the available data) as {LK attacks with Surrogate Learners} (LK-SL). 
This scenario can be denoted with $\vct \theta_{\rm LK-SL}=(\set D, \set X, \hat{\set M}, \hat{\vct w})$, even though the parameter vector $\hat{\vct w}$ may belong to a different vector space than that of the targeted learner.
Note that LK-SL attacks also include the case in which the attacker knows the learning algorithm, but she is not able to derive an optimal attack strategy against it (\eg, if the corresponding optimization problem is not tractable or difficult to solve), and thus uses a surrogate learning model to this end.
Experiments on the \emph{transferability} of attacks among learning algorithms, firstly demonstrated in~\cite{biggio13-ecml} and then in subsequent work on deep learners~\cite{papernot17-asiaccs}, fall under this category of attacks.

\subsection{Attacker's Capability} \label{sect:cap}

This characteristic is defined based on the \textbf{influence} that the attacker has on the input data, and on the presence of \textbf{data manipulation constraints}.

\myparagraph{Attack Influence} In supervised learning, the attack influence can be causative, if the attacker can influence both training and test data, or exploratory, if the attacker can only manipulate test data. These settings are more commonly referred to as \emph{poisoning} and \emph{evasion} attacks~\cite{barreno06-asiaccs,huang11,biggio14-tkde,biggio13-ecml,biggio12-icml,biggio15-icml,mei15-aaai}. 

\myparagraph{Data Manipulation Constraints} Another aspect related to the attacker's capability is the presence of constraints on the manipulation of input data, which is however strongly dependent on the given practical scenario. For example, if the attacker aims to evade a malware classification system, she should manipulate the exploitation code embedded in the malware sample without compromising its intrusive functionality. In the case of poisoning, the labels assigned to the training samples are not typically under the control of the attacker. She should thus consider additional constraints while manipulating the poisoning samples to have them labelled as desired. Typically, these constraints can be nevertheless accounted for in the definition of the optimal attack strategy. In particular, we characterize them by assuming that an initial set of attack samples $\set D_{c}$ is given, and that it is modified according to a space of possible modifications $\Phi(\set D_{c})$.

\subsection{Attack Strategy}
Given the attacker's knowledge $\vct \theta\in\Theta$ and a set of manipulated attack samples $\set D_{c}^{\prime} \in \Phi(\set D_{c})$, the attacker's goal can be characterized in terms of an objective function $\set A (\set D_{c}^{\prime}, \vct \theta) \in \mathbb R$ which evaluates how effective the attacks $\set D_{c}^{\prime}$ are.
The optimal attack strategy can be thus given as: 
\begin{eqnarray}
\set D_{c}^{\star} \in \argmax_{\set D_{c}^{\prime} \in \Phi(\set D_{c})}  \set A(\set D_{c}^{\prime}, \vct \theta) 
\label{eq:optim}
\end{eqnarray}
While this high-level formulation encompasses both evasion and poisoning attacks, in both binary and multiclass problems, in the remainder of this work we only focus on the definition of some poisoning attack scenarios.

\subsection{Poisoning Attack Scenarios} \label{ssec:AttackScenarios}

We focus here on two poisoning attack scenarios of interest for multiclass problems, noting that other attack scenarios can be derived in a similar manner.

\myparagraph{Error-Generic Poisoning Attacks} The most common scenario considered in previous work~\cite{biggio12-icml,biggio15-icml,mei15-aaai} considers poisoning two-class learning algorithms to cause a \emph{denial of service}.
This is an availability attack, and it could be targeted or indiscriminate, depending on whether it affects a specific system user or service, or any of them. In the multiclass case, it is thus natural to extend this scenario assuming that the attacker is not aiming to cause specific errors, but only \emph{generic} misclassifications.
As in~\cite{biggio12-icml,biggio15-icml,mei15-aaai}, this poisoning attack (as any other poisoning attack) requires solving a bilevel optimization, where the inner problem is the learning problem. This can be made explicit by rewriting Eq.~\eqref{eq:optim} as:
\begin{align}
  \label{eqObjective1} \set D_c^{\star} \in & \argmax_{\set D_{c}^{\prime} \in \Phi(\set D_{c})} && \set A(\set D_{c}^{\prime}, \vct \theta) = L (\hat{\set D}_{\rm val}, \hat{\vct w}) \, ,  \\
 \label{eqObjective2} &{\rm s.t.}   && \hat{\vct w} \in \argmin_{\vct w^{\prime} \in \set W} \set L(\hat{\set D}_{\rm tr} \cup \set D_{c}^{\prime}, \vct w^{\prime}) \, ,
\end{align}
where the surrogate data $\hat{\set D}$ available to the attacker is divided into two disjoint sets $\hat{\set D}_{\rm tr}$ and $\hat{\set D}_{\rm val}$. The former, along with the poisoning points $\set D_{c}^{\prime}$ is used to learn the surrogate model, while the latter is used to evaluate the impact of the poisoning samples on  untainted data, through the function $\set A(\set D_{c}^{\prime}, \vct \theta)$.
In this case, the function $\set A(\set D_{c}^{\prime}, \vct \theta)$ is simply defined in terms of a loss function $L (\hat{\set D}_{\rm val}, \hat{\vct w})$ that evaluates the performance of the (poisoned) surrogate model on $\hat{\set D}_{\rm val}$. The dependency of $\set A$ on $\set D_{c}^{\prime}$ is thus indirectly encoded through the parameters $\hat{\vct w}$ of the (poisoned) surrogate model.\footnote{Note that $\set A$ can also be directly dependent on $\set D_{c}^{\prime}$, as in the case of nonparametric models; \eg, in kernelized SVMs, when the poisoning points are support vectors~\cite{biggio12-icml}.}
Note that, since the learning algorithm (even if convex) may not exhibit a unique solution in the feasible set $\set W$, the outer problem has to be evaluated using the exact solution $\hat{\vct w}$ found by the inner optimization.
Worth remarking, this formulation encompasses all previously-proposed poisoning attacks against binary learners~\cite{biggio12-icml,biggio15-icml,mei15-aaai}, provided that the loss function $L$ is selected accordingly (\eg, using the hinge loss against SVMs~\cite{biggio12-icml}). In the multiclass case, one can use a multiclass loss function, like the log-loss with softmax activation, as done in our experiments.

\myparagraph{Error-Specific Poisoning Attacks} Here, we assume that the attacker's goal is to cause specific misclassifications -- a plausible scenario only for multiclass problems. This attack can cause an integrity or an availability violation, and it can also be targeted or indiscriminate, depending on the desired misclassifications. 
The poisoning problem remains that given by Eqs.~\eqref{eqObjective1}-\eqref{eqObjective2}, though the objective is defined as:
\begin{equation}
\set A(\set D_{c}^{\prime}, \vct \theta) = - L (\hat{\set D}^{\prime}_{\rm val}, \hat{\vct w}) \, ,
\label{eqObjectiveSpecific}
\end{equation}
where $\hat{\set D}^{\prime}_{\rm val}$ is a set that contains the same data as $\hat{\set D}_{\rm val}$, though with different labels, chosen by the attacker. These labels correspond to the desired misclassifications, and this is why there is a minus sign in front of $L$, \ie, the attacker effectively aims at \emph{minimizing} the loss on her desired set of labels. Note that, to implement an integrity violation or a targeted attack, some of these labels may actually be the same as the true labels (such that normal system operation is not compromised, or only specific system users are affected).

\section{Poisoning Attacks with Back-Gradient Optimization}
\label{sec:RMD}

In this section, we first discuss how the bilevel optimization given by Eqs.~\eqref{eqObjective1}-\eqref{eqObjective2} has been solved in previous work to develop gradient-based poisoning attacks~\cite{biggio12-icml,biggio15-icml,mei15-aaai,koh17-icml}. 
As we will see, these attacks can only be used against a limited class of learning algorithms, excluding neural networks and deep learning architectures, due to the inherent complexity of the procedure used to compute the required gradient.
To overcome this limitation, we exploit a recent technique called \emph{back-gradient optimization}~\cite{domke12-aistats,maclaurin15-icml}, which allows computing the gradient of interest in a more computationally-efficient and stabler manner.
Notably, this enables us to devise the first poisoning attack able to target neural networks and deep learning architectures (without using any surrogate model).

Before delving into the technical details, we make the same assumptions made in previous work~\cite{biggio12-icml,biggio15-icml,mei15-aaai} to reduce the complexity of Problem~\eqref{eqObjective1}-\eqref{eqObjective2}: $(i)$ we consider the optimization of one poisoning point at a time, denoted hereafter with $\vct x_{c}$; and $(ii)$ we assume that its label $y_{c}$ is initially chosen by the attacker, and kept fixed during the optimization. 
The poisoning problem can be thus simplified as: 
\begin{align}
\label{eqBilevelSingle1} \vct x_{c}^\star \in & \argmax_{\vct x_{c}^{\prime} \in \Phi(\{\vct x_{c}, y_{c}\})} &&
 \set A(\{\vct x_{c}^{\prime}, y_{c}\}, \vct \theta) = L (\hat{\set D}_{\rm val}, \hat{\vct w}) \, ,\\ \label{eqBilevelSingle2}
& {\rm s.t.} && \hat{\vct w} \in \argmin_{{\vct w}^{\prime} \in \set W} \ \set L( \vct x_c^{\prime}, {\vct w}^{\prime}) \, .
\end{align}
The function $\Phi$ imposes constraints on the manipulation of $\vct x_{c}$, \eg, upper and lower bounds on its manipulated values. These may also depend on $y_{c}$, \eg, to ensure that the poisoning sample is labelled as desired when updating the targeted classifier. 
Note also that, for notational simplicity, we only report $\vct x_{c}^{\prime}$ as the first argument of $\set L$ instead of $\hat{\set D}_{\rm tr} \cup \{\vct x_{c}^{\prime}, y_{c}\}$.

\myparagraph{Gradient-based Poisoning Attacks} 
We discuss here how Problem~\eqref{eqBilevelSingle1}-\eqref{eqBilevelSingle2} has been solved in previous work~\cite{biggio12-icml,biggio15-icml,mei15-aaai,koh17-icml}. For some classes of loss functions $L$ and learning objective functions $\set L$, this problem can be indeed solved through \emph{gradient ascent}. In particular, provided that the loss function $L$ is differentiable w.r.t.~$\vct w$ and $\vct x_{c}$, we can compute the gradient $\nabla_{{\vct x_c}} \set A$ using the chain rule: 
\begin{equation}
\nabla_{{\vct x_c}} \set A =  
\nabla_{\vct x_c} L + 
\diff{\hat{\vct w}}{\vct x_{c}}^\T
\nabla_{{\vct w}} L  \, ,
\label{eqGradient}
\end{equation}
where $L(\hat{\set D}_{\rm val}, \hat{\vct w})$ is evaluated on the parameters $\hat{\vct w}$ learned after training (including the poisoning point). The main difficulty here is computing $\diff{\hat{\vct w}}{\vct x_{c}}$, \ie, understanding how the solution of the learning algorithm varies w.r.t. the poisoning point.  
Under some regularity conditions, this can be done by replacing the inner learning problem with its stationarity (KKT) conditions.
For example, this holds if the learning problem $\set L$ is convex, which implies that all stationary points are global minima~\cite{pedregosa16-icml}. In fact, poisoning attacks have been developed so far only against learning algorithms with convex objectives~\cite{biggio12-icml,biggio15-icml,mei15-aaai,koh17-icml}.
The trick here is to replace the inner optimization with the implicit function $\nabla_{{\vct w}} \set L(\set D_{\rm tr} \cup \{ \vct x_c, y_c \}, \hat{\vct w}) = \vct 0$, corresponding to its KKT conditions. Then, assuming that it is differentiable w.r.t. $\vct x_{c}$, one yields the linear system $\nabla_{\vct x_c} \nabla_{{\vct w}} {\set L}  + \diff{\hat{\vct w}}{\vct x_{c}}^\T \nabla^{2}_{{\vct w}} {\set L} = \vct 0$.
If $\nabla^{2}_{{\vct w}} {\set L}$ is not singular, we can solve this system w.r.t. $\diff{\hat{\vct w}}{\vct x_{c}}$, and substitute its expression in Eq.~\eqref{eqGradient}, yielding:
\begin{equation}
\nabla_{\vct x_c}{\set A} = \nabla_{\vct x_c} L -( \nabla_{\vct x_c} \nabla_{{\vct w}} {\set L}) (\nabla_{{\vct w}}^2 {\set L})^{-1} \nabla_{{\vct w}} {L} \, .
\label{eqImplicit}
\end{equation} 
This gradient is then iteratively used to update the poisoning point through gradient ascent, as shown in Algorithm~\ref{alg:poisoning}.\footnote{Note that Algorithm~\ref{alg:poisoning} can be exploited to optimize multiple poisoning points too. As in~\cite{biggio15-icml}, the idea is to perform several passes over the set of poisoning samples, using Algorithm~\ref{alg:poisoning} to optimize each poisoning point at a time, while keeping the other points fixed. Line searches can also be exploited to reduce complexity.}
Recall that the projection operator $\Pi_{\Phi}$ is used to map the current poisoning point onto the feasible set $\Phi$ (cf. Eqs.~\ref{eqBilevelSingle1}-\ref{eqBilevelSingle2}).

\begin{algorithm}[tb]
  \caption{Poisoning Attack Algorithm}
  \label{alg:poisoning}
\begin{flushleft}
  \textbf{Input:} $\hat{\set D}_{\rm tr}$, $\hat{\set D}_{\rm val}$, $\set L$, $L$, the initial poisoning point $ \vct x_{c}^{(0)}$, its label $y_{c}$, the learning rate $\eta$, a small positive constant $\varepsilon$.  
\end{flushleft}  
  \begin{algorithmic}[1]
  \STATE{$i \leftarrow 0$ (iteration counter)}
		\REPEAT
		\STATE{$\hat{\vct w} \in \argmin_{{\vct w}^{\prime}} \ \set L( \vct x_c^{(i)}, {\vct w}^{\prime})$ (train learning algorithm)}	
		\STATE{$\vct x_{c}^{(i+1)} \leftarrow \Pi_{\Phi} \left ({\vct x_c}^{(i)} + 
			\eta  \nabla_{{\vct x_c}} {\set A}(\{\vct x_c^{(i)}, y_{c}\}) \right )$}
		\STATE{$i \leftarrow i + 1$}
		\UNTIL{${\set A}(\{\vct x_c^{(i)}, y_{c}\}) - {\set A}(\{\vct x_c^{(i-1)}, y_{c}\}) < \varepsilon$}
  \end{algorithmic}
  \begin{flushleft}
  \textbf{Output:} the final poisoning point $\vct x_{c} \leftarrow \vct x_{c}^{(i)}$
  \end{flushleft}
  
\end{algorithm}

This is the state-of-the-art approach used to implement current poisoning attacks~\cite{biggio12-icml,biggio15-icml,mei15-aaai,koh17-icml}. The problem here is that computing and inverting $\nabla_{{\vct w}}^2 {\set L}$ scales in time as ${\mathcal O}(p^3)$ and in memory as ${\mathcal O}(p^2)$, being $p$ the cardinality of $\vct w$. Moreover, Eq.~\eqref{eqImplicit} requires solving one linear system per parameter. These aspects make it prohibitive to assess the effectiveness of poisoning attacks in a variety of practical settings. 

To mitigate these issues, as suggested in~\cite{do2008,domke12-aistats,maclaurin15-icml,koh17-icml}, one can apply conjugate gradient descent to solve a simpler linear system, obtained by a trivial re-organization of the terms in the second part of Eq.~\eqref{eqImplicit}. In particular, one can set $(\nabla_{{\vct w}}^2 {\set L}) \ {\bf v} = \nabla_{{\vct w}} L$, and compute $\nabla_{{\bf x_c}}{\set A} = \nabla_{\vct x_c} L - \nabla_{{\bf x_c}} \nabla_{{\vct w}} {\set L} \ {\bf v}$. The computation of the matrices $\nabla_{{\vct x_c}} \nabla_{{\vct w}} {\set L}$ and $\nabla_{{\vct w}}^2 {\set L}$ can also be avoided using Hessian-vector products~\cite{pearlmutter}: 
\begin{align}
\nonumber
(\nabla_{\vct x_c} \nabla_{{\vct w}} {\set L}) \ {\bf z} & = \lim_{h \rightarrow 0} \frac{1}{h} \left( \nabla_{\vct x_{c}} \set L \left(\vct x_c^{\prime},\hat{\vct w} + h{\bf z} \right) - \nabla_{\vct x_c} \set L \left(\vct x_c^{\prime},\hat{\vct w} \right) \right) \, , \\
\nonumber
(\nabla_{{\vct w}}\nabla_{{\vct w}} {\set L}) \ {\bf z} & = \lim_{h \rightarrow 0} \frac{1}{h} \left( \nabla_{\vct w} \set L \left(\vct x_c^{\prime},\hat{\vct w} + h{\bf z} \right) - \nabla_{\vct w} \set L \left(\vct x_c^{\prime},\hat{\vct w} \right) \right) \, .
\end{align}
Although this approach allows poisoning learning algorithms more efficiently w.r.t. previous work~\cite{biggio12-icml,biggio15-icml,mei15-aaai}, it still requires the inner learning problem to be solved exactly. From a practical perspective, this means that the KKT conditions have to be met with satisfying numerical accuracy.
However, as these problems are always solved to a finite accuracy, 
it may happen that the gradient $\nabla_{\vct x_{c}} \set A$ is not sufficiently precise, especially if convergence thresholds are too loose~\cite{domke12-aistats,maclaurin15-icml}.

It is thus clear that such an approach can not be used, in practice, to poison learning algorithms like neural networks and deep learning architectures, as it may not only be difficult to derive proper stationarity conditions involving all parameters, but also as it may be too computationally demanding to train such learning algorithms with sufficient precision to correctly compute the gradient $\nabla_{\vct x_{c}} \set A$.

\begin{algorithm}[tb]
  \caption{Gradient Descent}
  \label{alg:grad} 
  \begin{flushleft}
  \textbf{Input:} initial parameters ${\vct w}_0$, learning rate ${\eta}$, $\hat{\set D}_{\rm tr}$, $\set L$.
   \end{flushleft}    
  
  \begin{algorithmic}[1]
  	\FOR{$t=0,\ldots, T-1$}
  	\STATE {${\bf g}_t = \nabla_{{\vct w}} {\set L} (\hat{\set D}_{\rm tr}, {\vct w}_t)$}
  	\STATE {${\vct w}_{t + 1} \leftarrow {\vct w}_t - \eta \ {\bf g}_t$}
  	\ENDFOR
  \end{algorithmic}
  \begin{flushleft}
  \textbf{Output:} trained parameters ${\vct w}_T$
  \end{flushleft}   
\end{algorithm}

\begin{algorithm}[tb]
  \caption{Back-gradient Descent}
  \label{alg:reverse}
  \begin{flushleft}
  \textbf{Input:} trained parameters ${\vct w}_T$, learning rate $\eta$, $\hat{\set D}_{\rm tr}$, $\hat{\set D}_{\rm val}$, poisoning point $\vct x_{c}^{\prime}$, $y_{c}$, loss function $L$, learner's objective $\set L$.\\
  initialize $d\vct x_c \leftarrow {\bf 0}$, $d{\vct w} \leftarrow \nabla_{\vct w} L(\hat{\set D}_{\rm val}, {\vct w}_T)$ 
  \end{flushleft}     
  
  \begin{algorithmic}[1]
        \FOR{$t= T,\ldots,1$}
        \STATE{$d{\vct x_c} \leftarrow d \vct x_c^{\prime} - \eta \, d{\vct w} \nabla_{\vct  x_c} 
        \nabla_{\vct w} {\set L} (\vct x_c^{\prime}, \vct w_t)$}
        \STATE{$d{\vct w} \leftarrow d{\vct w} - \eta \, d{\vct w} \nabla_{\vct w} 
        \nabla_{\vct w} {\set L} (\vct x_c^{\prime}, \vct w_t)$}
        \STATE{${\bf g}_{t-1} = \nabla_{{\vct w}_t} {\set L} (\vct x_c^{\prime}, \vct w_t)$}
        \STATE{${\vct w}_{t-1} = {\vct w}_{t} + \alpha {\bf g}_{t-1}$}
      \ENDFOR
  \end{algorithmic}
  \begin{flushleft}
  \textbf{Output:} $\nabla_{{\vct x_c}} \set A = \nabla_{\vct x_c} L + d{\vct x_c}$
  \end{flushleft}
\end{algorithm}

\begin{figure*}
	\centering
	\includegraphics[width=0.95\textwidth]{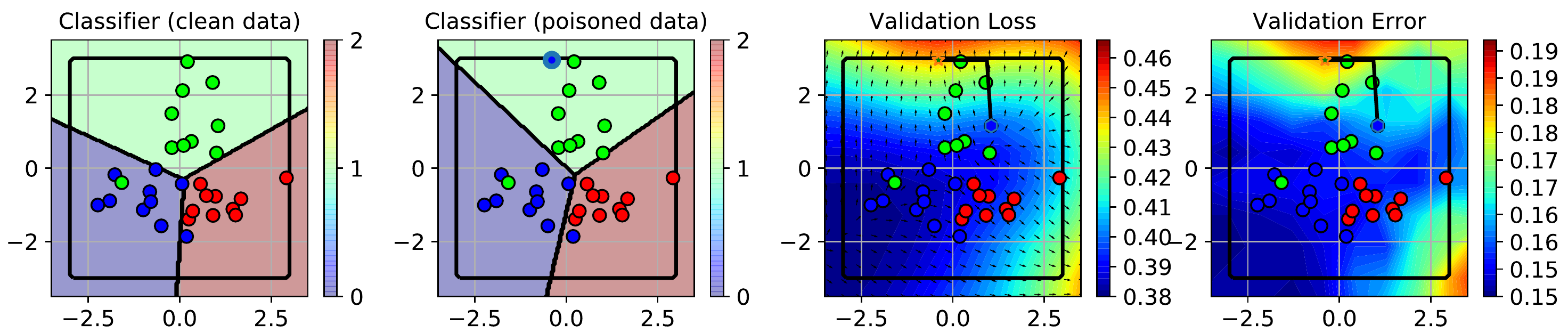}
	\includegraphics[width=0.95\textwidth]{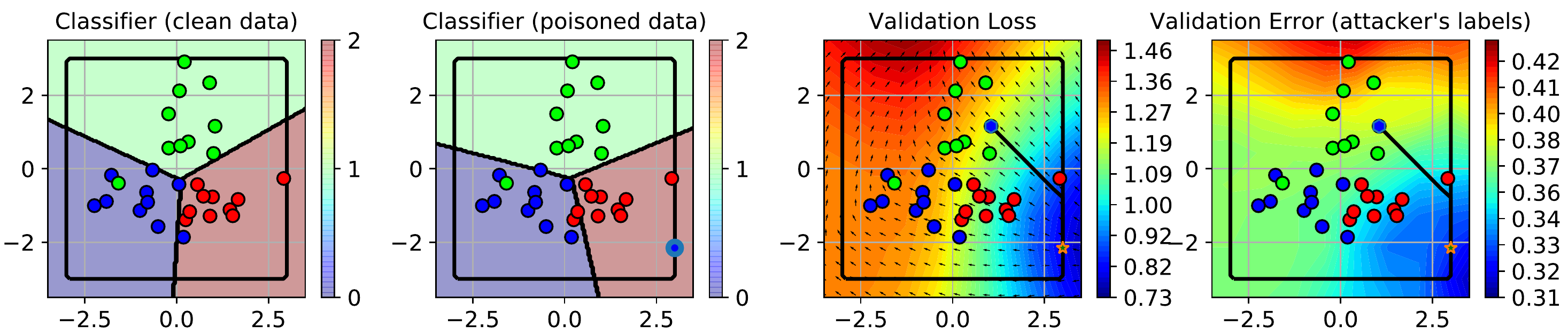}
	\caption{Error-generic ({top row}) and error-specific ({bottom row}) poisoning attacks on a three-class synthetic dataset, against a multiclass logistic classifier. In the error-specific case, the attacker aims to have red points misclassified as blue, while preserving the labels of the other points. We report the decision regions on the clean ({first column}) and on the poisoned ({second column}) data, in which we only add a poisoning point labelled as blue (highlighted with a blue circle). The validation loss $L (\hat{\set D}_{\rm val}, \hat{\vct w})$ and $L (\hat{\set D}^{\prime}_{\rm val}, \hat{\vct w})$, respectively maximized in error-generic and minimized in error-specific attacks, is shown in colors, as a function of the attack point $\vct x_{c}$ ({third column}), along with the corresponding back-gradients (shown as arrows), and the path followed while optimizing $\vct x_{c}$. To show that the logistic loss used to estimate $L$ provides a good approximation of the true error, we also report the validation error measured with the zero-one loss on the same data ({fourth column}).}
	\label{fig:2d-example}	
\end{figure*}

\myparagraph{Poisoning with Back-gradient Optimization} In this work, we overcome this limitation by exploiting \emph{back-gradient optimization}~\cite{domke12-aistats,maclaurin15-icml}. 
This technique has been first exploited 
in the context of energy-based models and hyperparameter optimization, to solve bilevel optimization problems similar to the poisoning problem discussed before.
The underlying idea of this approach is to replace the inner optimization with a set of iterations performed by the learning algorithm to update the parameters $\vct w$, provided that such updates are \emph{smooth}, as in the case of gradient-based learning algorithms.
According to \cite{domke12-aistats}, this technique allows to compute the desired gradients in the outer problem using the parameters $\vct w_{T}$ obtained from an incomplete optimization of the inner problem (after $T$ iterations). This represent a significant computational improvement compared to traditional gradient-based approaches, since it only requires a reduced number of training iterations for the learning algorithm. This is especially important in large neural networks and deep learning algorithms, where the computational cost per iteration can be high.
Then, assuming that the inner optimization runs for $T$ iterations, the idea is to exploit reverse-mode differentiation, or \emph{back-propagation}, to compute the gradient of the outer objective. 
However, using back-propagation in a na\"ive manner would not work for this class of problems,
as it requires storing the whole set of parameter updates $\vct w_{1}, \ldots, \vct w_{T}$ performed during training, along with the forward derivatives. These are indeed the elements required to compute the gradient of the outer objective with a \emph{backward} pass (we refer the reader to~\cite{maclaurin15-icml} for more details). 
This process can be extremely memory-demanding if the learning algorithm runs for a large number of iterations $T$, and especially if the number of parameters $\vct w$ is large (as in deep networks).
Therefore, to avoid storing the whole training trajectory $\vct w_{1}, \ldots, \vct w_{T}$ and the required forward derivatives, \citet{domke12-aistats} and \citet{maclaurin15-icml} proposed to compute them directly during the backward pass, by \emph{reversing} the steps followed by the learning algorithm to update them. Computing $\vct w_{T}, \ldots, \vct w_{1}$ in reverse order w.r.t. the forward step is clearly feasible only if the learning procedure can be exactly traced backwards.
Nevertheless, this happens to be feasible for a large variety of gradient-based procedures, including gradient descent with fixed step size, and stochastic gradient descent with momentum.

In this work, we leverage back-gradient descent to compute $\nabla_{{\vct x_c}} {\set A}$ (Algorithm \ref{alg:reverse}) by reversing a standard gradient-descent procedure with fixed step size that runs for a truncated training of the learning algorithm to $T$ iterations (Algorithm~\ref{alg:grad}). Notably, lines 2-3 in Algorithm~\ref{alg:reverse} can be efficiently computed with Hessian-vector products, as discussed before. 
We exploit this algorithm to compute the gradient $\nabla_{{\vct x_c}} {\set A}$ in line 4 of our poisoning attack algorithm (Algorithm~\ref{alg:poisoning}). In this case, line 3 of Algorithm~\ref{alg:poisoning} is replaced with the incomplete optimization of the learning algorithm, truncated to $T$ iterations. 
Note finally that, as in \cite{domke12-aistats,maclaurin15-icml}, the time complexity of our back-gradient descent is ${\mathcal O}(T)$. This drastically reduces the complexity of the computation of the outer gradient, making it feasible to evaluate the effectiveness of poisoning attacks also against large neural networks and deep learning algorithms. Moreover, this outer gradient can be accurately estimated from a truncated optimization of the inner problem with a reduced number of iterations. This allows for a tractable computation of the poisoning points in Algorithm~\ref{alg:poisoning}, since training the learning algorithm at each iteration can be prohibitive, especially for deep networks.

We conclude this section by noting that, in the case of error-specific poisoning attacks (Sect.~\ref{ssec:AttackScenarios}), the outer objective in Problem~\eqref{eqBilevelSingle1}-\eqref{eqBilevelSingle2} is $-L(\hat{\set D}_{val}^{\prime}, \hat{\vct w})$.
This can be regarded as a minimization problem, and it thus suffices to modify line 4 in Algorithm~\ref{alg:poisoning} to update the poisoning point along the opposite direction. 
We clarify this in Fig.~\ref{fig:2d-example}, where we also discuss the different effect of error-generic and error-specific poisoning attacks in a multiclass setting.

\section{Experimental Analysis}\label{sec:Experiments}

In this section, we first evaluate the effectiveness of the back-gradient poisoning attacks described in Sect.~\ref{sec:RMD} on spam and malware detection tasks. In these cases, we also assess whether poisoning samples can be \emph{transferred} across different learning algorithms.
We then investigate the impact of error-generic and error-specific poisoning attacks in the well-known multiclass problem of handwritten digit recognition. In this case, we also report the first proof-of-concept adversarial training examples computed by poisoning a convolutional neural network in an \emph{end-to-end} manner (\ie, not just using a surrogate model trained on the deep features, as in~\cite{koh17-icml}).

\subsection{Spam and Malware Detection} 

We consider here two distinct datasets, respectively representing a spam email classification problem (\texttt{Spambase}) and a malware detection task (\texttt{Ransomware}).
The \texttt{Spambase} data~\cite{uci} consists of a collection of $4,601$ emails, including $1,813$ spam emails.
Each email is encoded as a feature vector consisting of $54$ binary features, each denoting the presence or absence of a given word in the email.
The \texttt{Ransomware} data~\cite{sgandurra} consists of $530$ ransomware samples and $549$ benign applications. 
Ransomware is a very recent kind of malware which encrypts the data on the infected machine, and requires the victim to pay a ransom to obtain the decryption key.
This dataset has $400$ binary features accounting for different sets of actions, API invocations, and modifications in the file system and registry keys during the execution of the software.

We consider the following leaning algorithms: $(i)$ Multi-Layer Perceptrons (MLPs) with one hidden layer consisting of $10$ neurons; $(ii)$ Logistic Regression (LR); and $(iii)$ Adaline (ADA). 
For MLPs, we have used hyperbolic tangent activation functions for the neurons in the hidden layer, and softmax activations in the output layer. Moreover, for MLPs and LR, we use the cross-entropy (or log-loss) as the loss function, while we use the mean squared error for ADA. 

We assume here that the attacker aims to cause a denial of service, and thus runs a poisoning \emph{availability} attack whose goal is simply to maximize the classification error.
Accordingly, we run Algorithm~\ref{alg:poisoning} injecting up to $20$ poisoning points in the training data. We initialize the poisoning points by cloning training points and flipping their label. 
We set the number of iterations $T$ for obtaining stable back-gradients to $200$, $100$, and $80$, respectively for MLPs, LR and ADA.
We further consider two distinct settings: PK attacks, in which the attacker is assumed to have full knowledge of the attacked system (for a worst-case performance assessment); and LK-SL attacks, in which she knows everything except for the learning algorithm, and thus uses a surrogate learner $\hat{\set M}$. This scenario, as discussed in Sect.~\ref{sect:knowledge}, is useful to assess the \emph{transferability} property of the attack samples.
To the best of our knowledge, this has been demonstrated in \cite{biggio13-ecml,papernot17-asiaccs} for evasion attacks (\ie, adversarial \emph{test} examples) but never for poisoning attacks (\ie, adversarial \emph{training} examples). To this end, we optimize the poisoning samples using alternatively MLPs, LR or ADA as the surrogate learner, and then evaluate the impact of the corresponding attacks against the other two algorithms.

\begin{figure}
	\centering
	\includegraphics[width=0.3\textwidth]{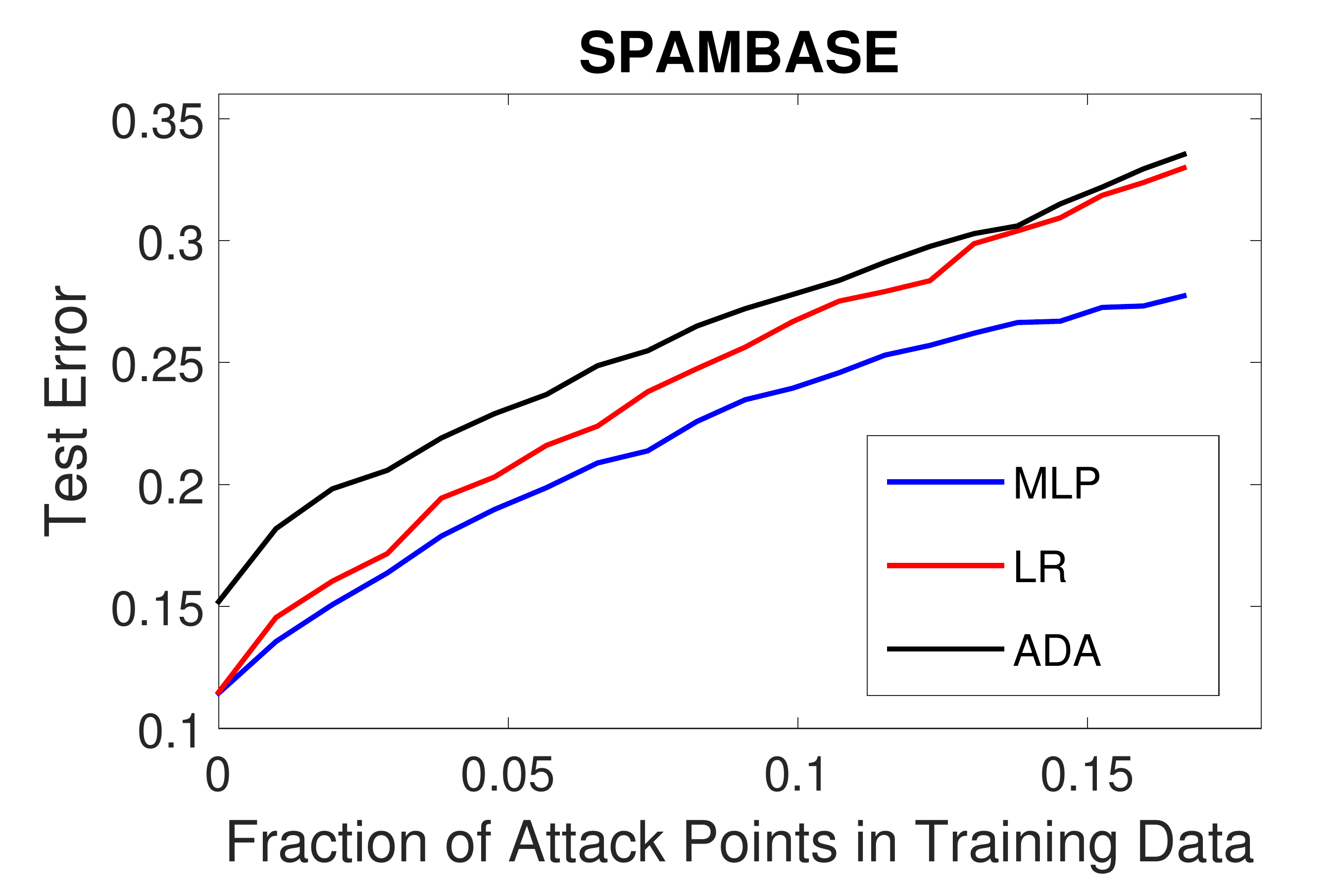}
	\includegraphics[width=0.3\textwidth]{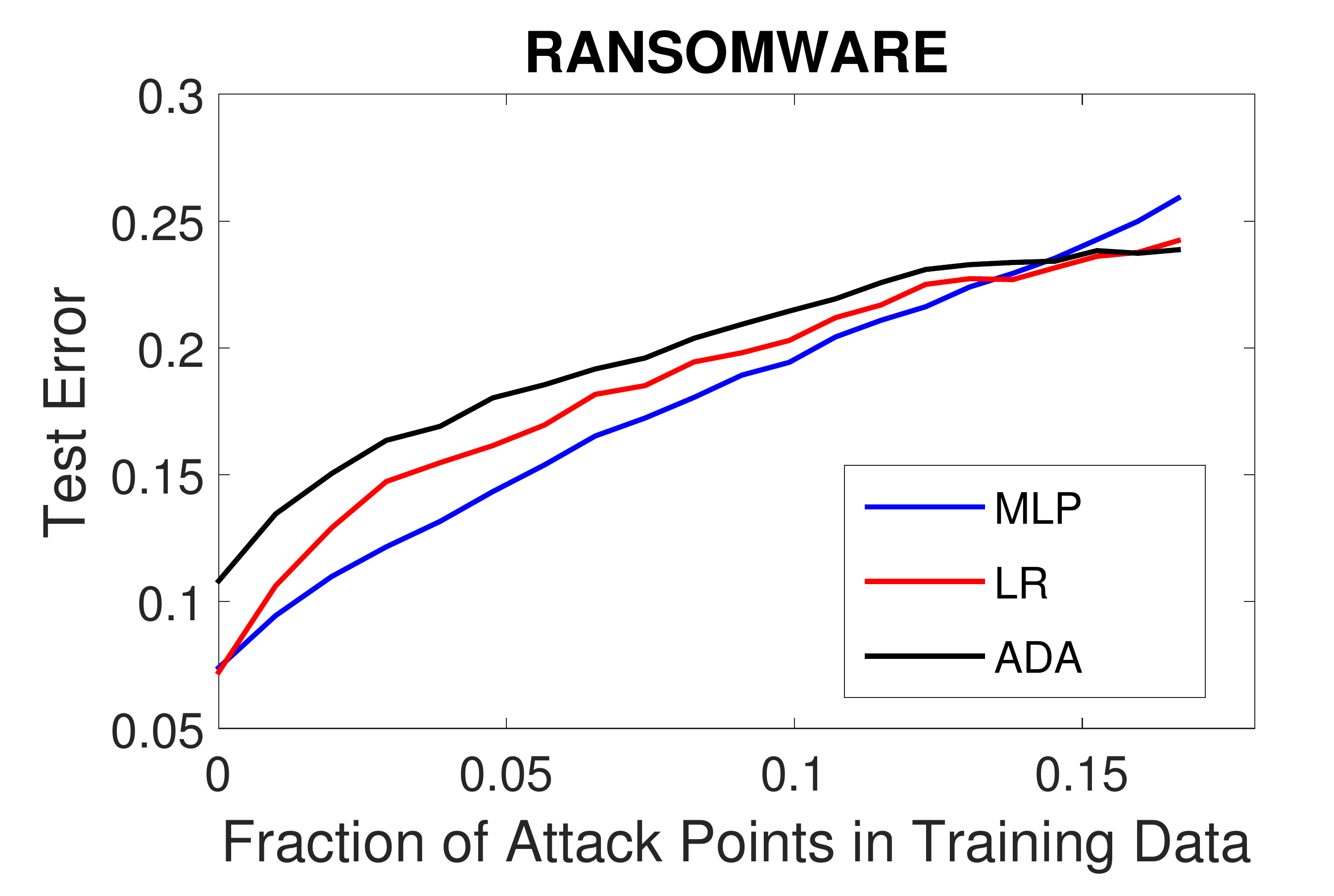}
	\caption{Results for PK poisoning attacks.}\vspace{-10pt}
	\label{fig:results-PK-spam-malware}	
\end{figure}

\begin{figure*}
	\centering
	\includegraphics[width=0.3\textwidth]{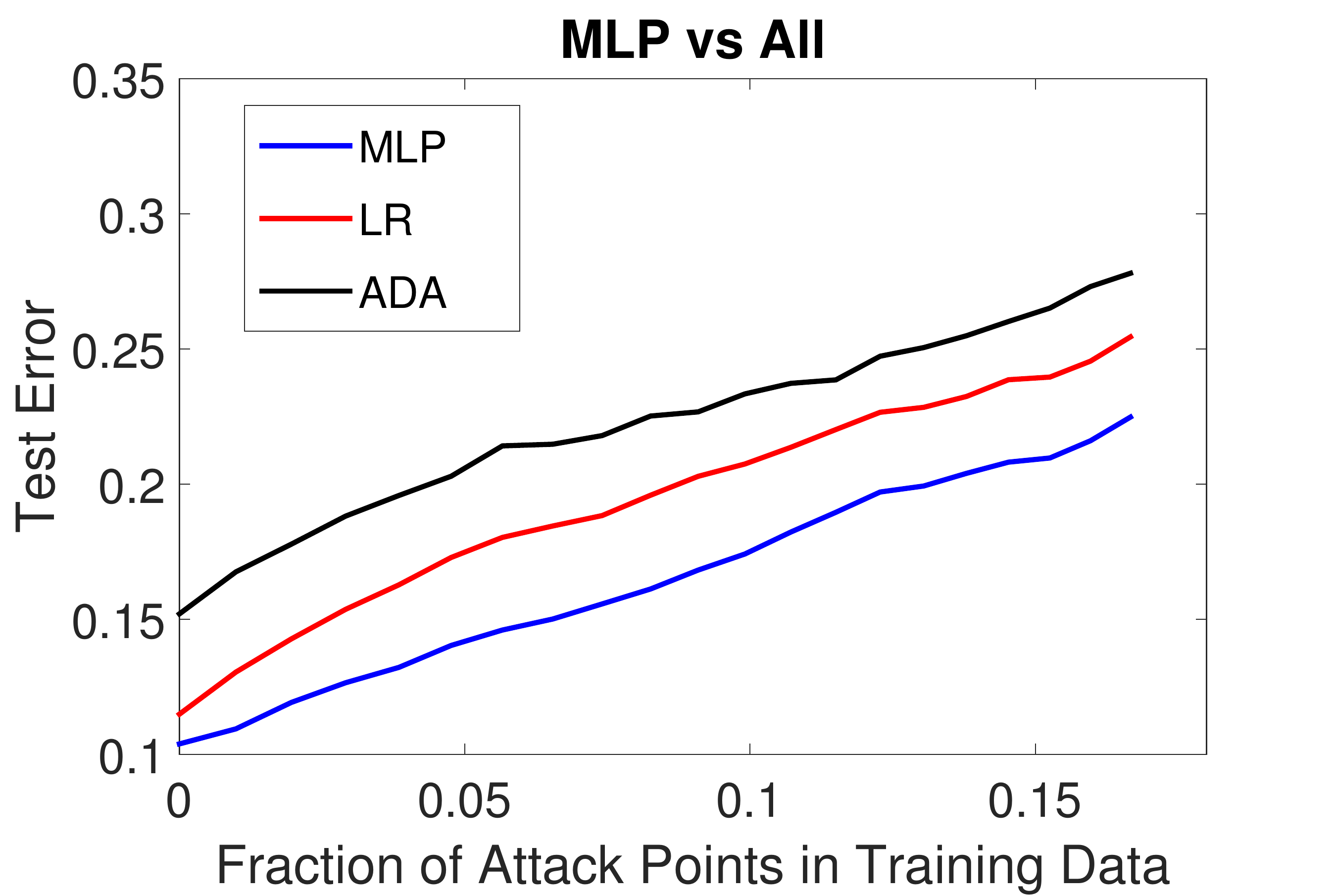}
	\includegraphics[width=0.3\textwidth]{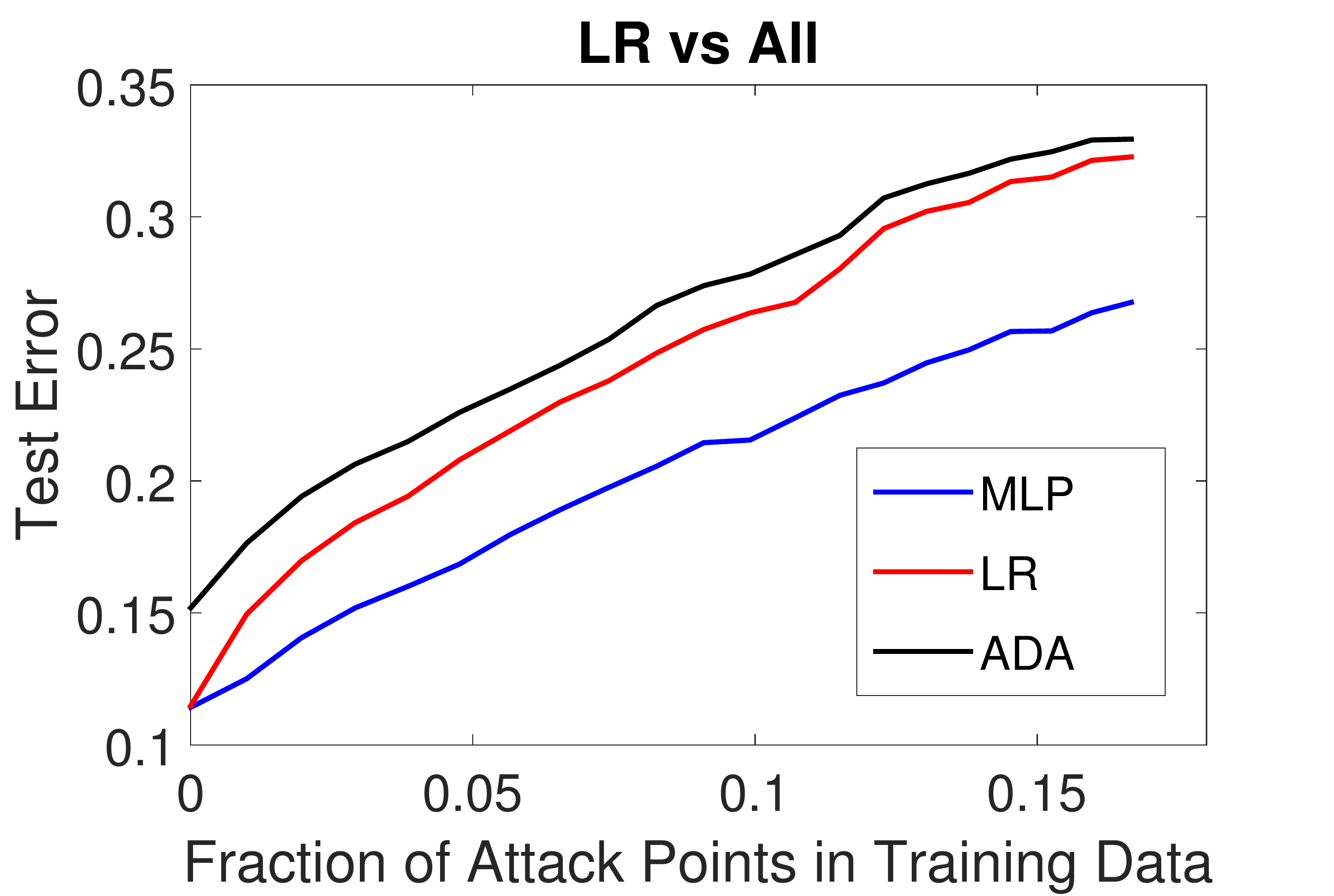}
	\includegraphics[width=0.3\textwidth]{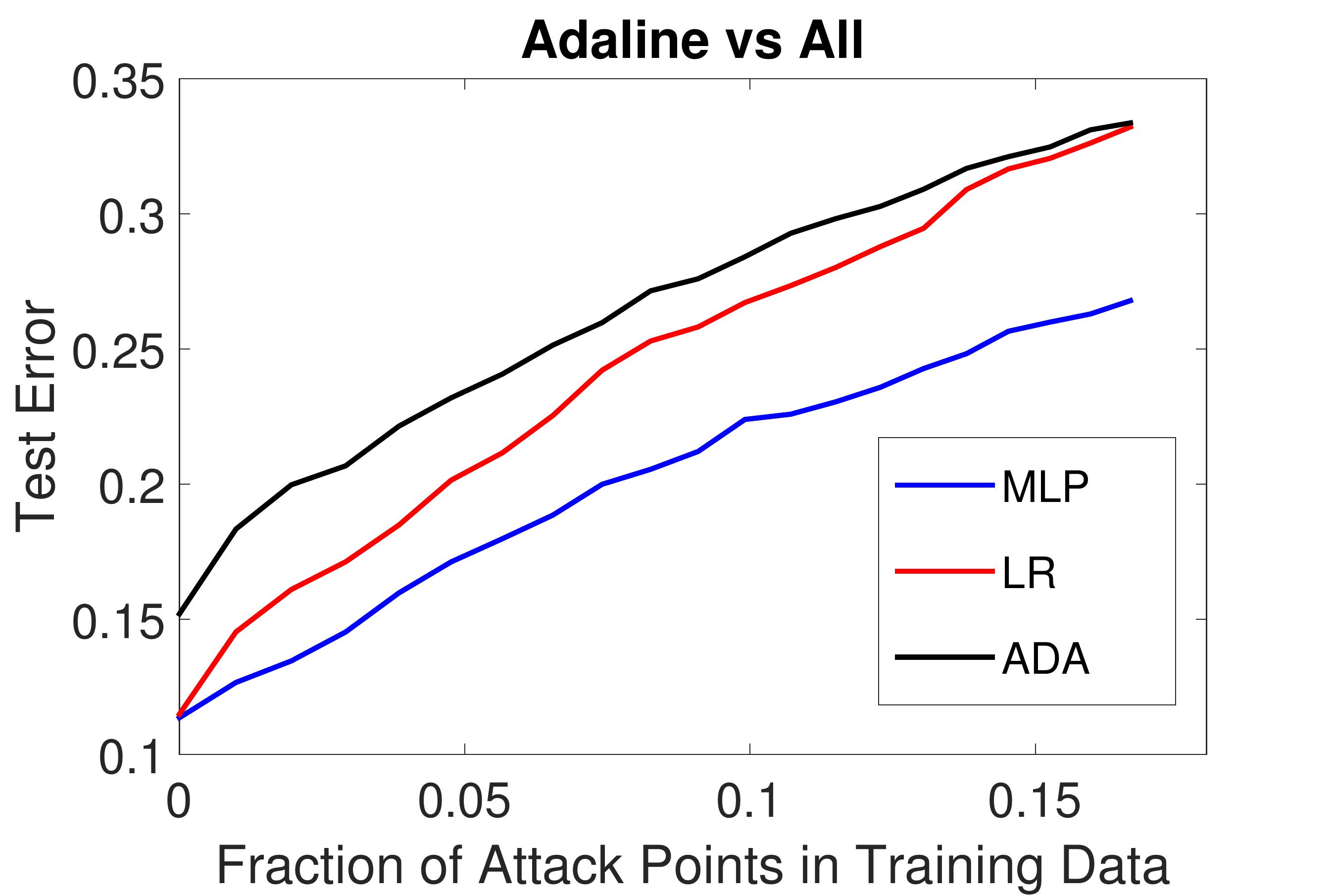}\\
	\includegraphics[width=0.3\textwidth]{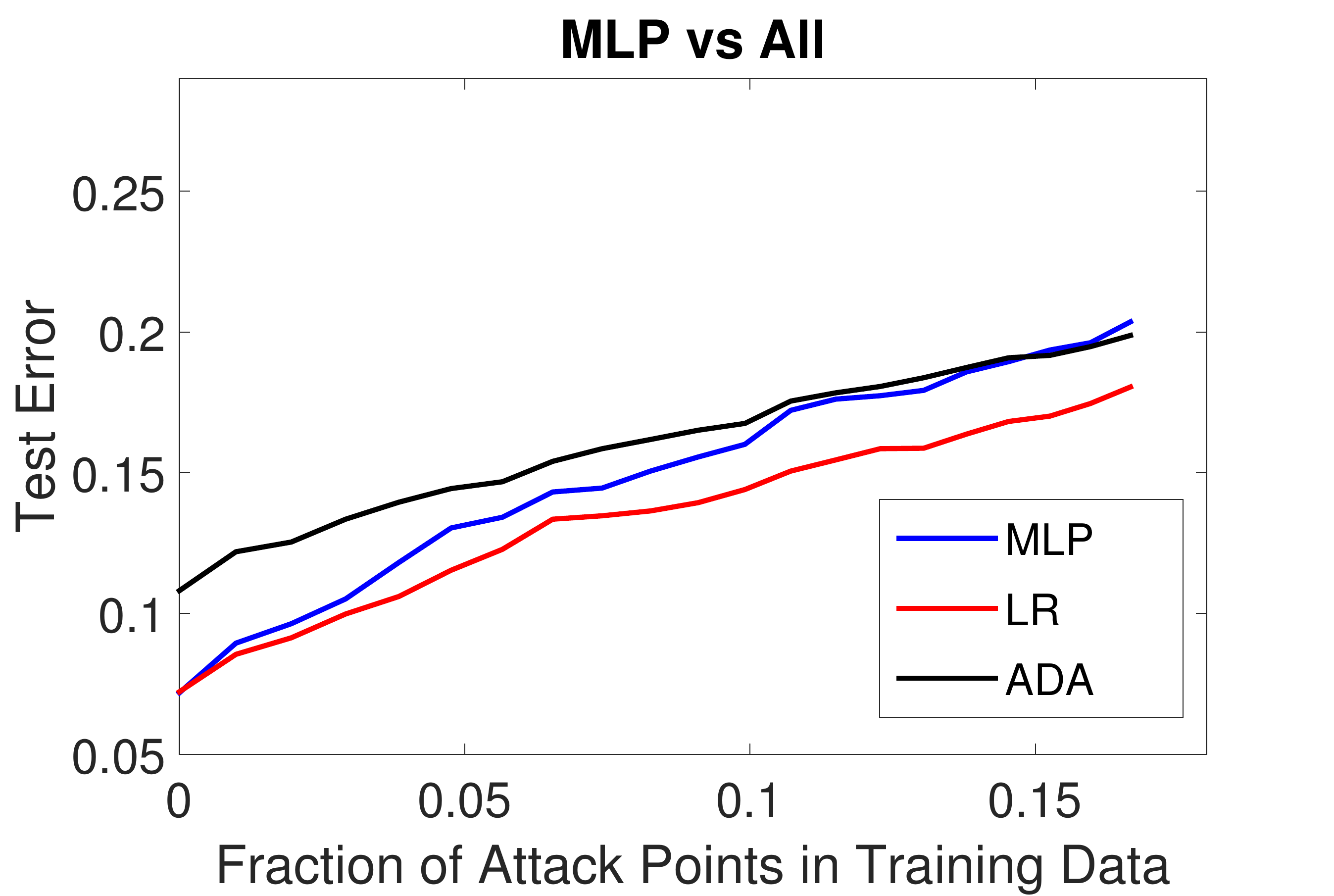}
	\includegraphics[width=0.3\textwidth]{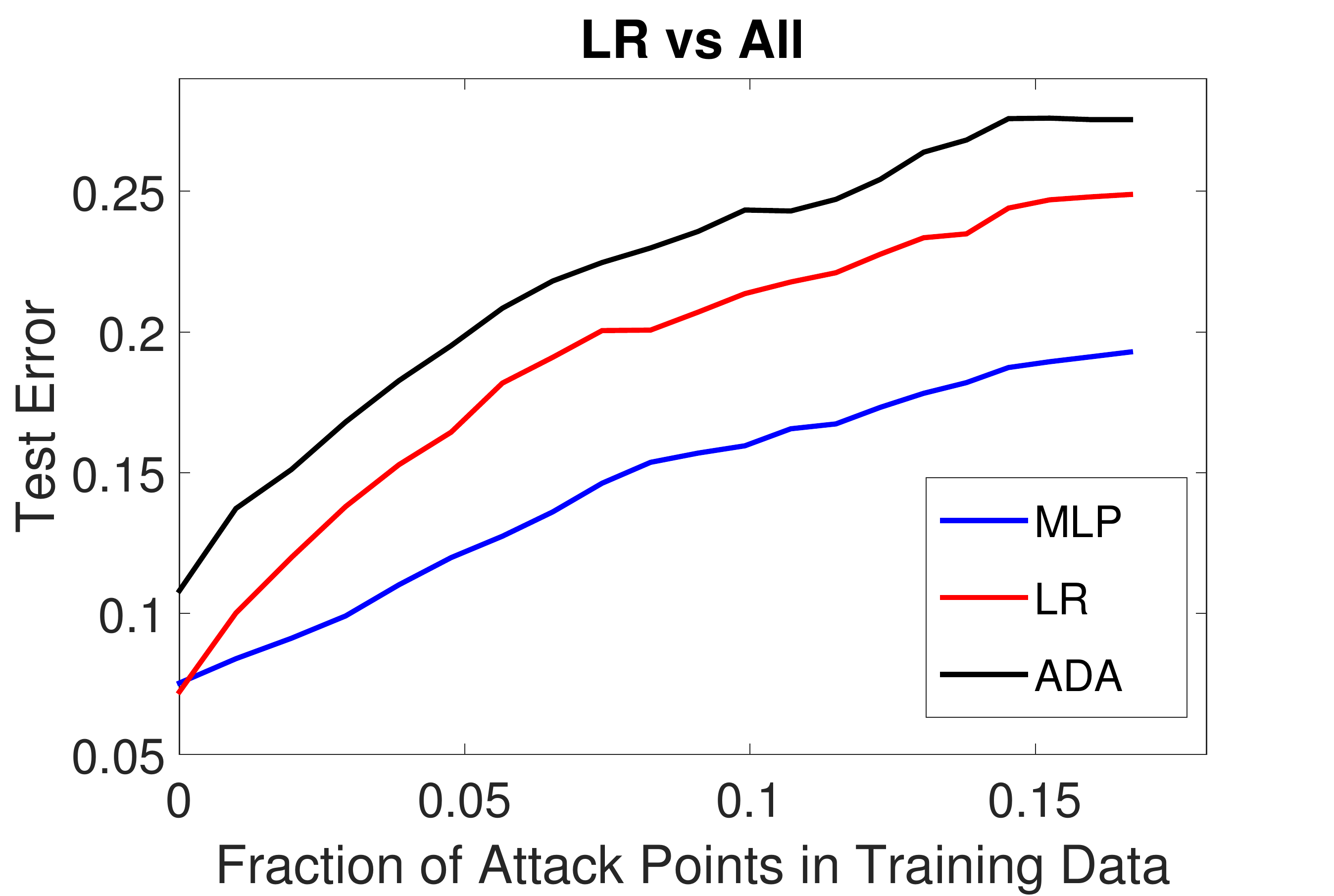}
	\includegraphics[width=0.3\textwidth]{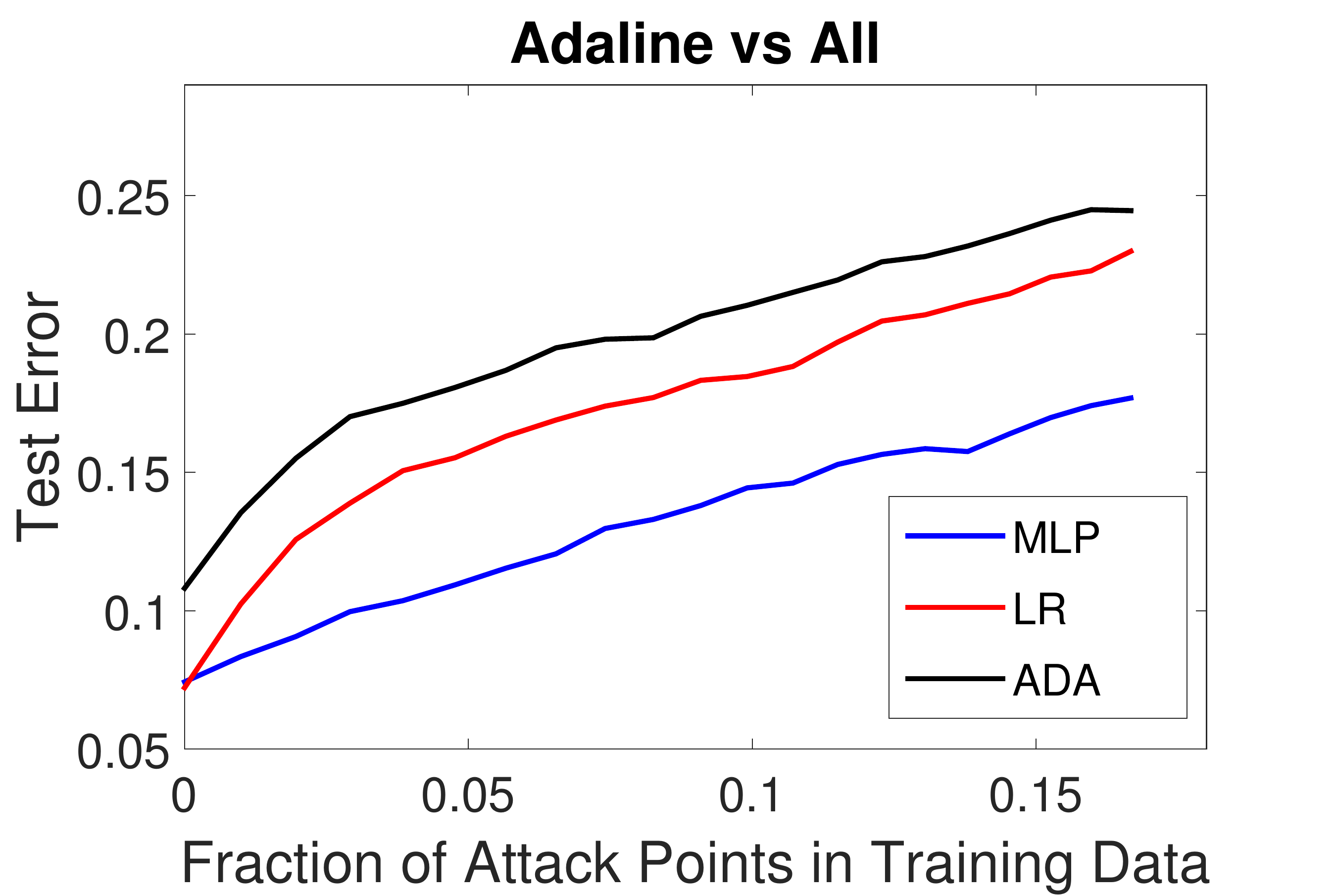}
	\caption{Results for LK-SL poisoning attacks (transferability of poisoning samples) on Spambase (top row) and Ransomware (bottom row).}\vspace{-5pt}
	\label{fig:results-LK-spam-malware}	
\end{figure*}

The experimental results, shown in Figs.~\ref{fig:results-PK-spam-malware}-\ref{fig:results-LK-spam-malware}, are averaged on $10$ independent random data splits. In each split, we use $100$ samples for training and $400$ for validation, \ie, to respectively construct $\set D_{\rm tr}$ and $\set D_{\rm val}$. Recall indeed that in both PK and LK-SL settings, the attacker has perfect knowledge of the training set used to learn the true (attacked) model, \ie, $\hat{\set D}_{\rm tr} = \set D_{\rm tr}$.
The remaining samples are used for testing, \ie, to assess the classification error under poisoning.\footnote{Note indeed that the validation error only provides a biased estimate of the true classification error, as it is used by the attacker to optimize the poisoning points~\cite{biggio12-icml}.}

We can observe from Fig.~\ref{fig:results-PK-spam-malware} that PK poisoning attacks can significantly compromise the performance of all the considered classifiers.
In particular, on \texttt{Spambase}, they cause the classification error of ADA and LR to increase up to $30\%$ even if the attacker only controls $15\%$ of the training data. Although the MLP is more resilient to poisoning than these linear classifiers, its classification error also increases significantly, up to $25\%$, which is not tolerable in several practical settings. The results for PK attacks on \texttt{Ransomware} are similar, although the MLP seems as vulnerable as ADA and LR in this case.

\myparagraph{Transferability of Poisoning Samples} Regarding LK-SL poisoning attacks, we can observe from Fig.~\ref{fig:results-LK-spam-malware} that the attack points generated using a linear classifier (either ADA or LR) as the surrogate model have a very similar impact on the other linear classifier. In contrast, the poisoning points crafted with these linear algorithms have a lower impact against the MLP, although its performance is still noticeably affected.
When the MLP is used as the surrogate model, instead, the performance degradation of the other algorithms is similar. However, the impact of these attacks is much lower. 
To summarize, our results show that the attack points can be effectively transferred across linear algorithms and also have a noticeable impact on (nonlinear) neural networks. In contrast, transferring poisoning samples from nonlinear to linear models seems to be less effective.

\subsection{Handwritten Digit Recognition}

We consider here the problem of handwritten digit recognition, which involves $10$ classes (each corresponding to a digit, from $0$ to $9$), using the MNIST data~\cite{lecun98}. 
Each digit image consists of $28 \times 28 = 784$ pixels, ranging from $0$ to $255$ (images are in grayscale). We divide each pixel value by $255$ and use it as a feature. We evaluate the effect of error-generic and error-specific poisoning strategies against a multiclass LR classifier using softmax activation and the log-loss as the loss function.

\myparagraph{Error-generic attack} In this case, the attacker aims to maximize the classification error regardless of the resulting kinds of error, as described in Sect.~\ref{ssec:AttackScenarios}.
This is thus an \emph{availability} attack, aimed to cause a denial of service.
We generate $10$ independent random splits using $1000$ samples for training, $1000$ for validation, and $8000$ for testing. To compute the back-gradients $\nabla_{\vct x_{c}} \set A$ required by our poisoning attack, we use $T=60$ iterations. We initialize the poisoning points by cloning randomly-chosen training points and changing their label at random
In addition, we compare our poisoning attack strategy here against 
a label-flip attack in which the attack points are drawn from the validation set and their labels are flipped at random. In both cases, we inject up to $60$ attack points into the training set.

\begin{figure*}
	\includegraphics[width=0.32\textwidth]{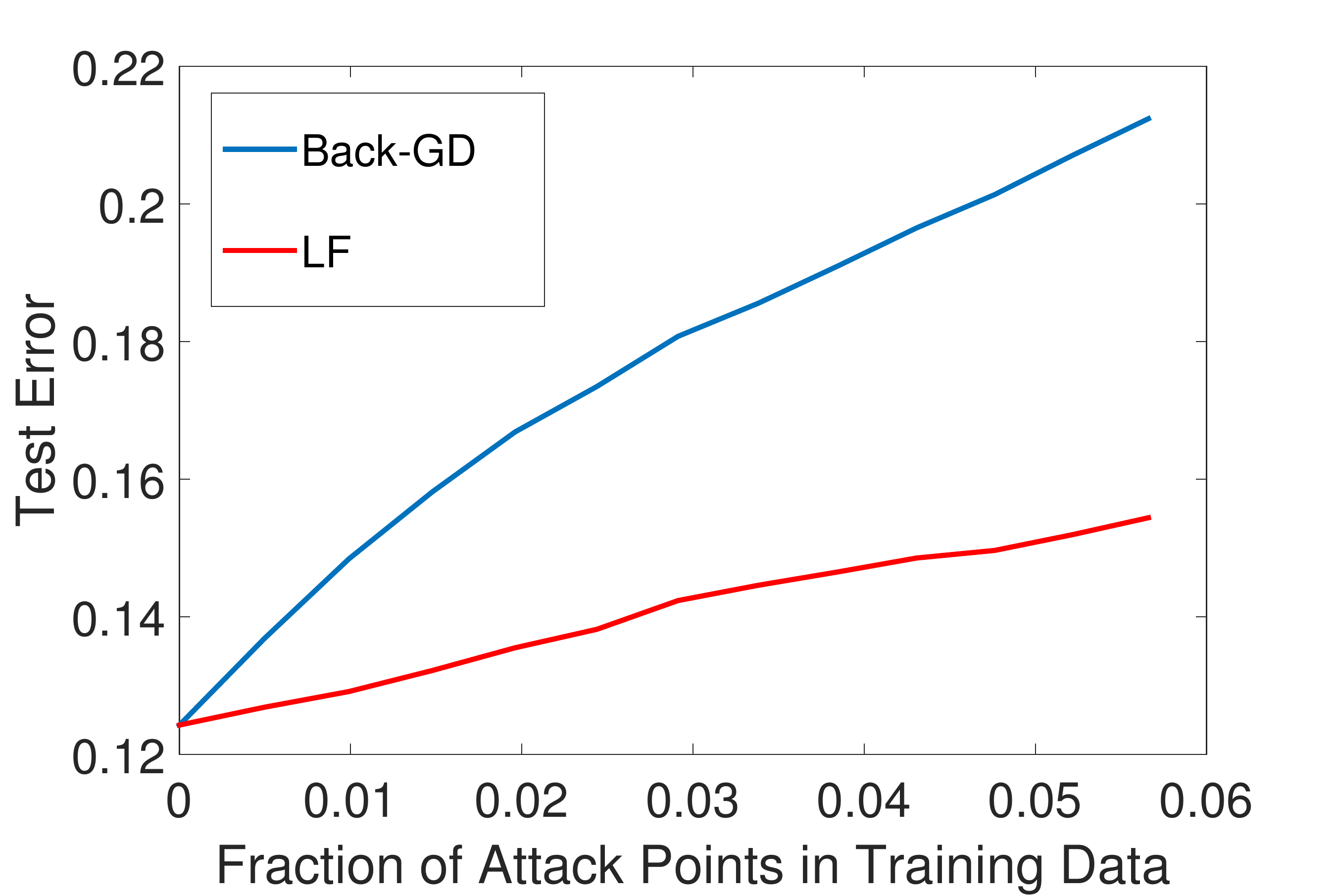}
	\includegraphics[width=0.32\textwidth]{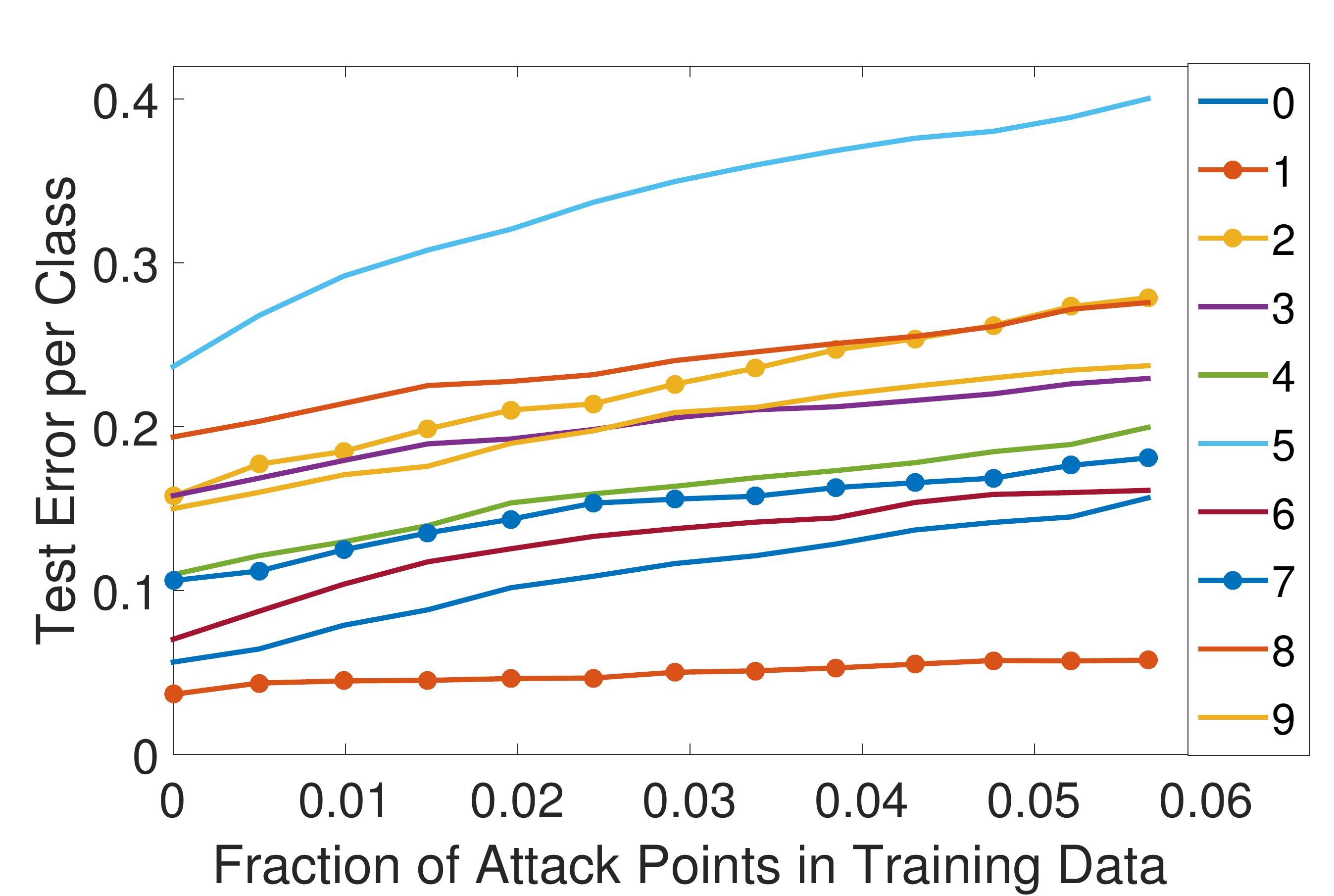}
	\includegraphics[width=0.32\textwidth]{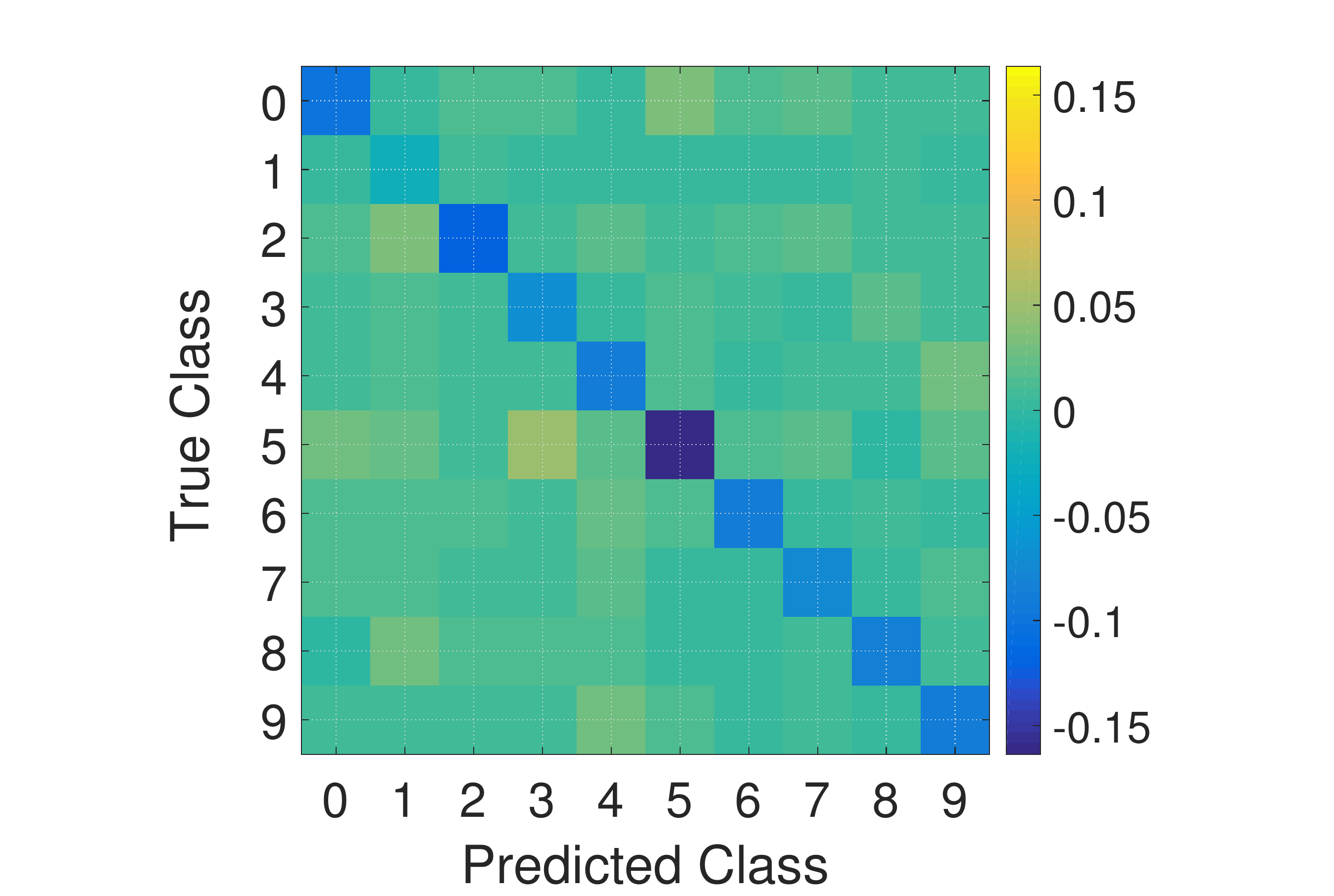}\\
	\includegraphics[width=0.32\textwidth]{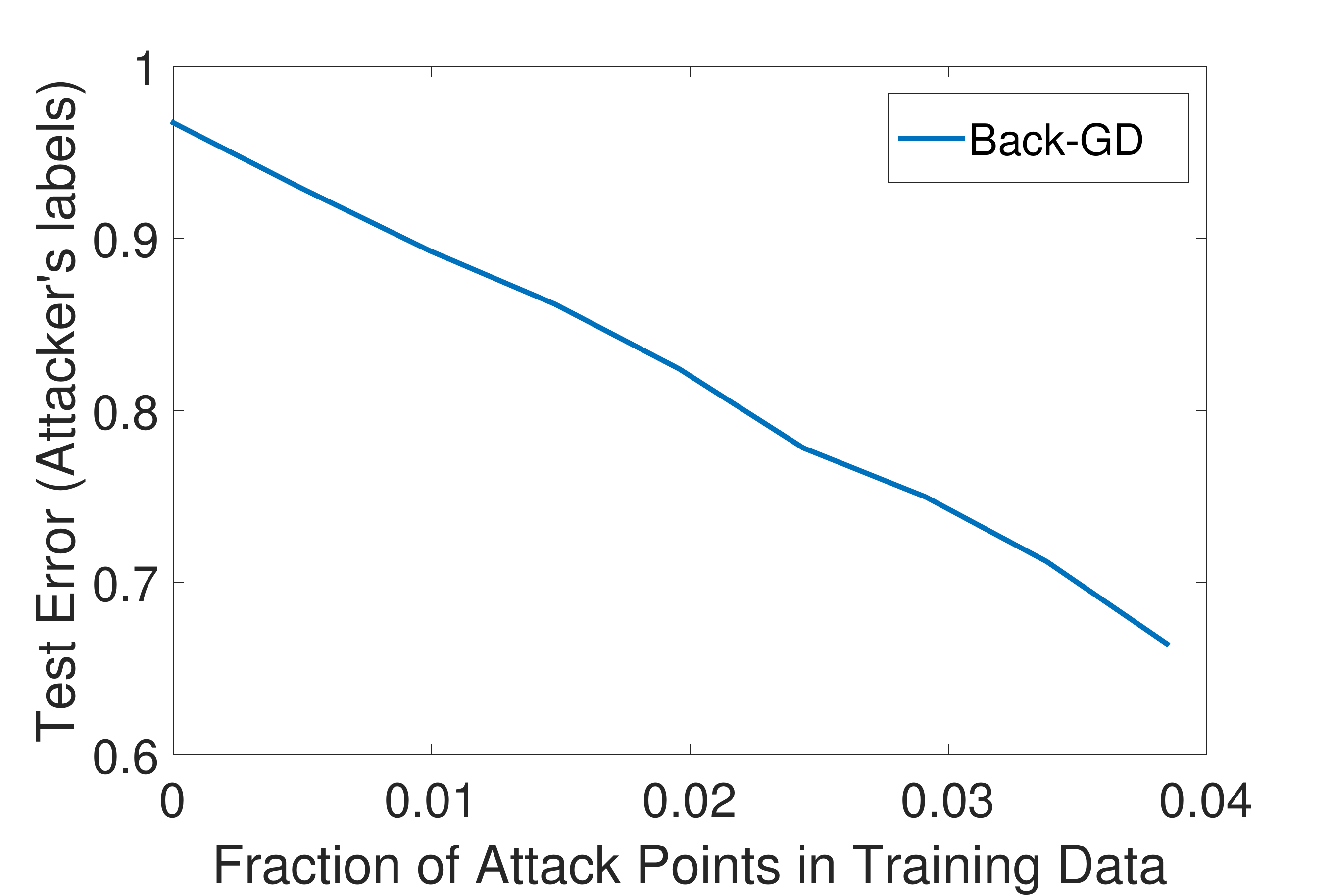}
	\includegraphics[width=0.32\textwidth]{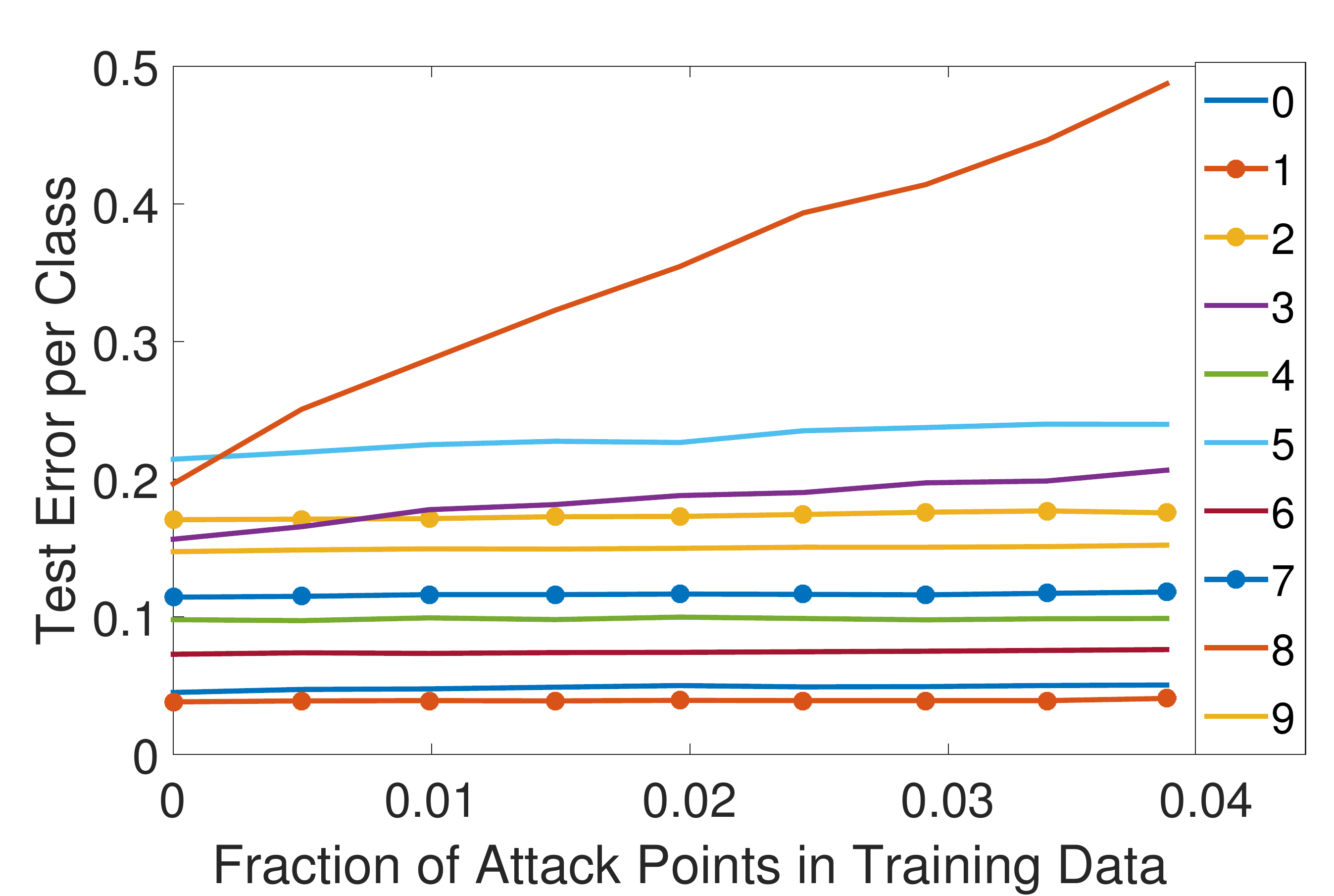}
	\includegraphics[width=0.32\textwidth]{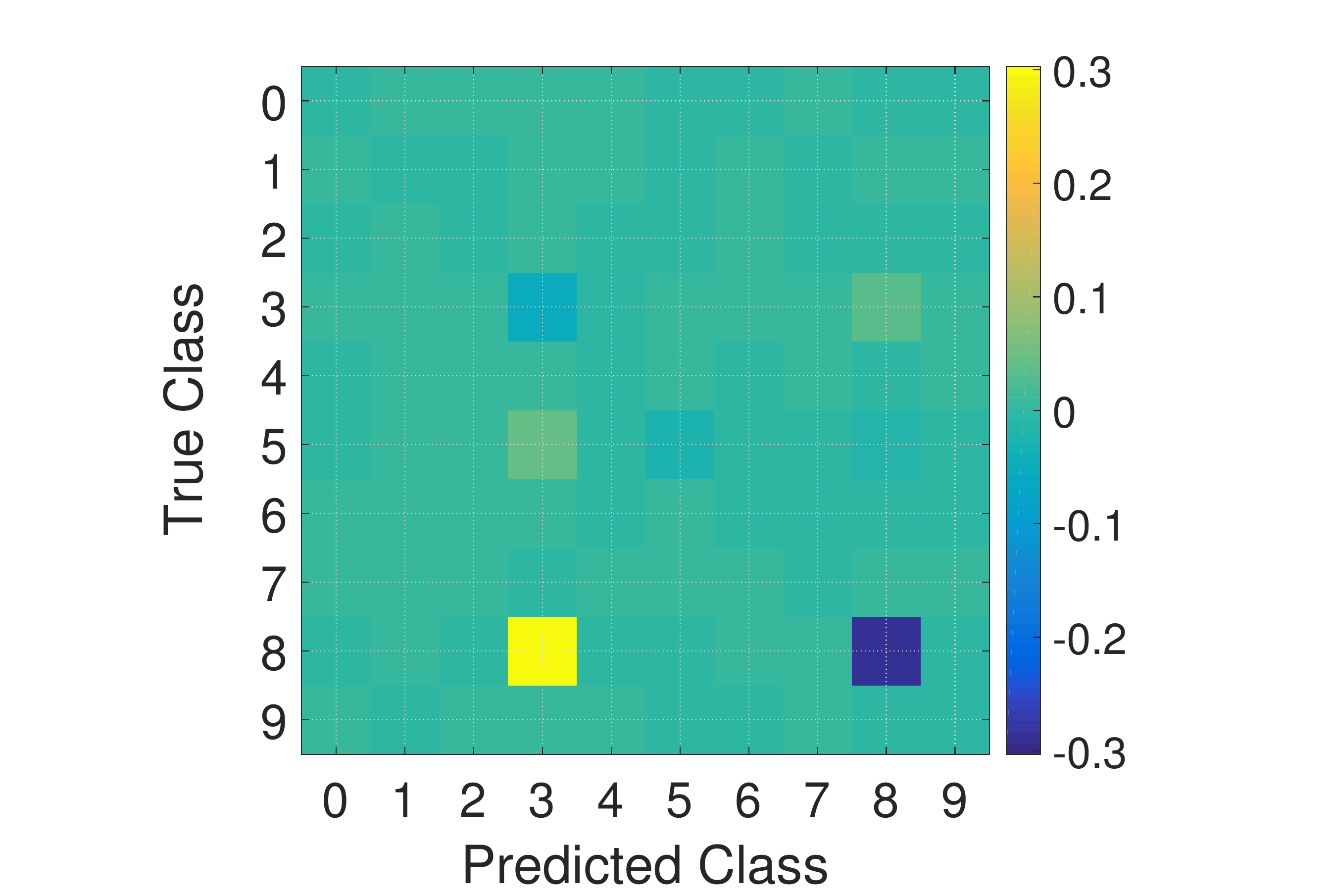}
\caption{Error-generic (top row) and error-specific (bottom row) poisoning against multiclass LR on the MNIST data. In the first column, we report the test error (which, for error-specific poisoning attacks, is computed using the attacker's labels instead of the true labels, and so it decreases while approaching the attacker's goal). In the second column, we report the error per class, \ie, the probability of misclassifying a digit given that it belongs to the class reported in the legend. In the third column, we report the difference between the confusion matrix obtained under poisoning (after injecting the maximum number of poisoning samples) and that obtained in the absence of attack, to highlight how the errors affect each class.}\vspace{-5pt}
\label{fig:mnist}
\end{figure*}

The results are shown in Fig.~\ref{fig:mnist} (top row). Note first that our error-generic poisoning attack almost doubles the classification error in the absence of poisoning, with less than $6\%$ of poisoning points. It is also much more effective than random label flips and, as expected, it causes a similar increase of the classification error over all classes (although some classes are easier to poison, like digit $5$).
This is even more evident from the difference between the confusion matrix obtained under $6\%$ poisoning and that obtained in the absence of attack.

\myparagraph{Error-specific attack} Here, we assume that the attacker aims to misclassify $8$s as $3$s, while not having any preference regarding the classification of the other digits. 
This can be thus regarded as an \emph{availability} attack, targeted to cause the misclassification of a specific set of samples.
We generate $10$ independent random splits with $1000$ training samples, $4000$ samples for validation, and $5000$ samples for testing.
Recall that the goal of the attacker in this scenario is described by Eq.~\eqref{eqObjectiveSpecific}.
In particular, she aims at minimizing $L(\hat{\set D}^{\prime}_{val}, \hat{\vct w})$, where the samples in the validation set $\hat{\set D}^{\prime}_{val}$ are re-labelled according to the attacker's goal. 
Here, the validation set thus only consists of digits of class $8$ labelled as $3$.
We set $T=60$ to compute the back-gradients used in our poisoning attack, and inject up to $40$ poisoning points into the training set. We initialize the poisoning points  by cloning randomly-chosen samples from the classes $3$ and $8$ in the training set, and flipping their label from $3$ to $8$, or vice-versa. We consider only these two classes here as they are the only two actively involved in the attack.

The results are shown in Fig.~\ref{fig:mnist} (bottom row). We can observe that only the classification error rate for digit $8$ is significantly affected, as expected. 
In particular, it is clear from the difference of the confusion matrix obtained under poisoning and the one obtained in the absence of attack that most of the $8$s are misclassified as $3$s.
After adding less than $4\%$ of poisoning points, in fact, the error rate for digit $8$ increases approximately from $20\%$ to $50\%$. Note that, as a side effect, the error rate of digit $3$ also slightly increases, though not to a significant extent.

\begin{figure}[t]
	\includegraphics[width=0.15\textwidth]{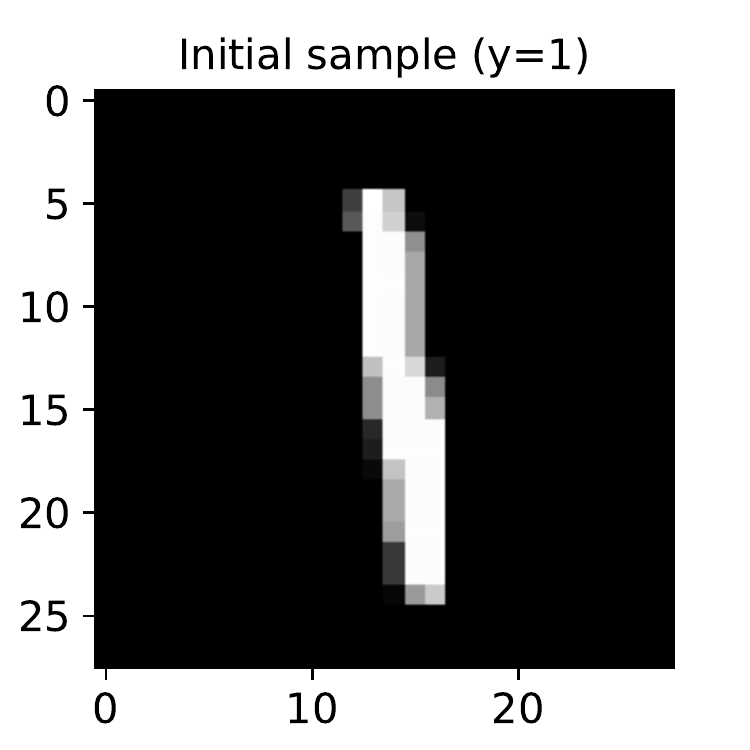}
	\includegraphics[width=0.15\textwidth]{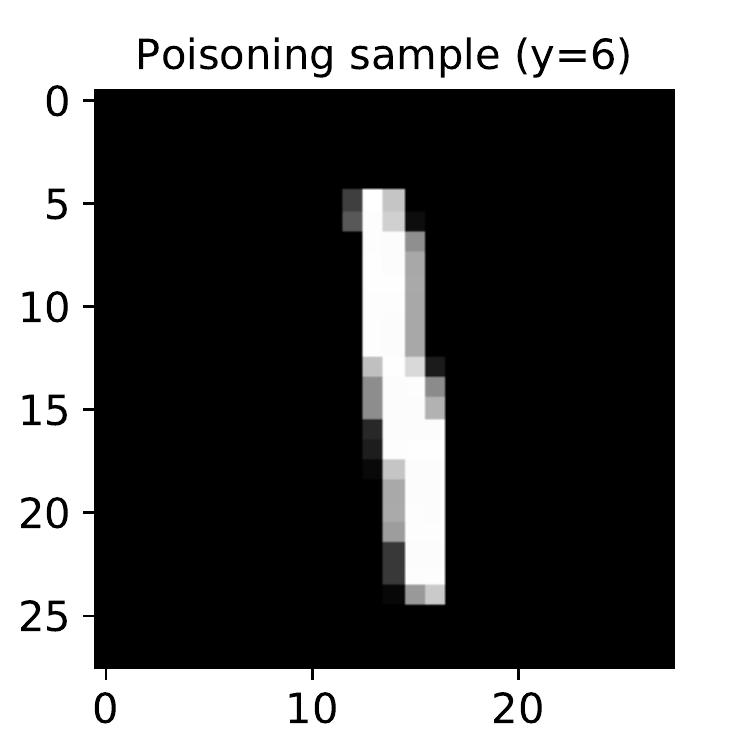}
	\includegraphics[width=0.15\textwidth]{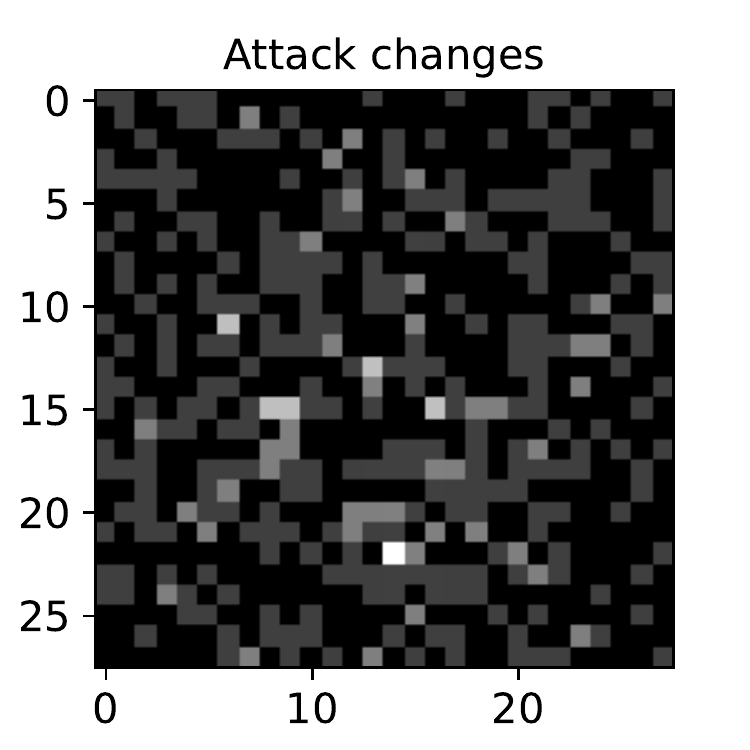}
	\includegraphics[width=0.15\textwidth]{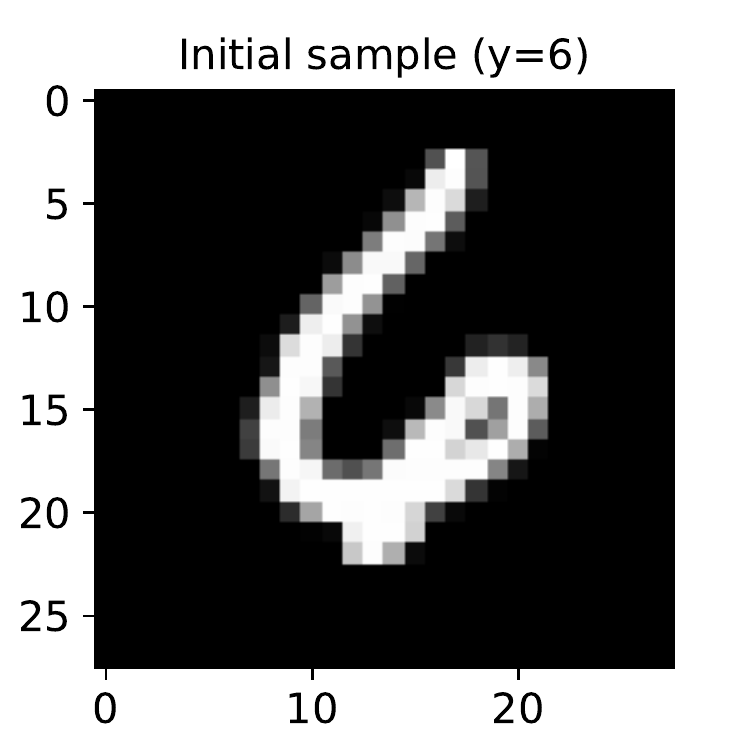}
	\includegraphics[width=0.15\textwidth]{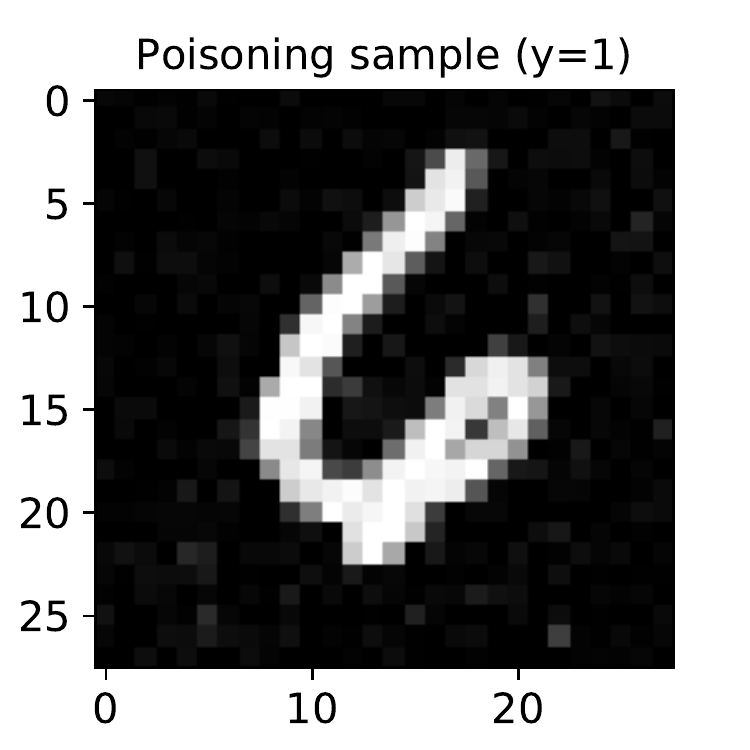}
	\includegraphics[width=0.15\textwidth]{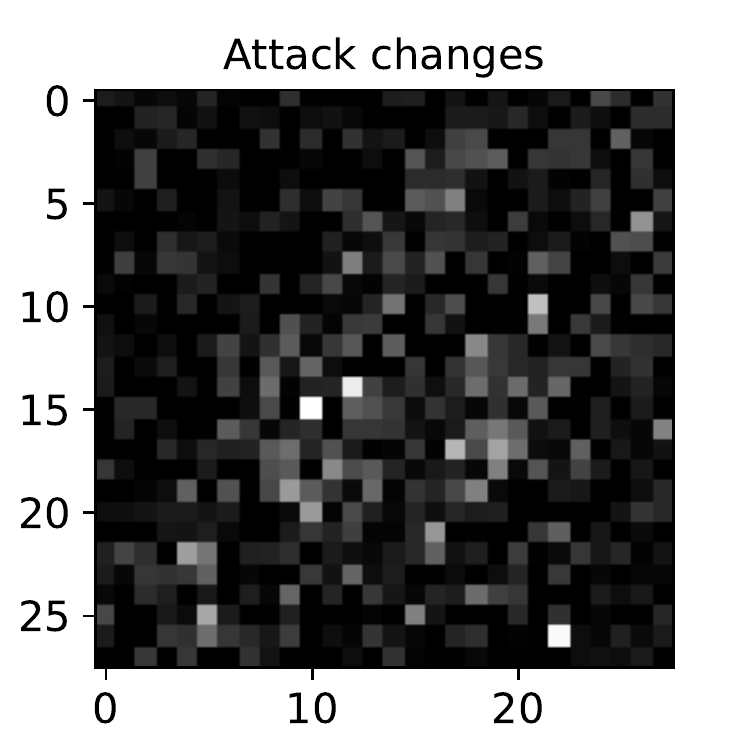}
	\includegraphics[width=0.15\textwidth]{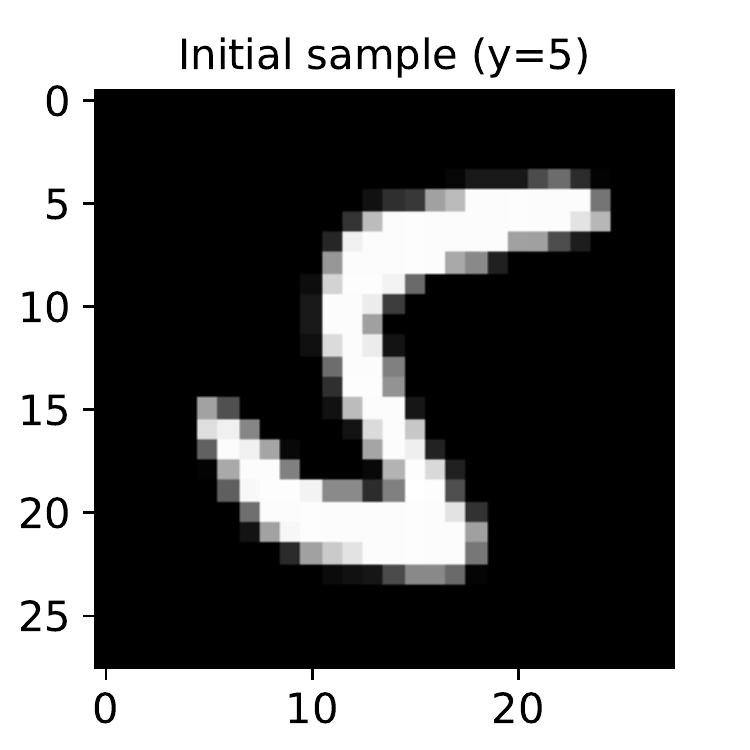}
	\includegraphics[width=0.15\textwidth]{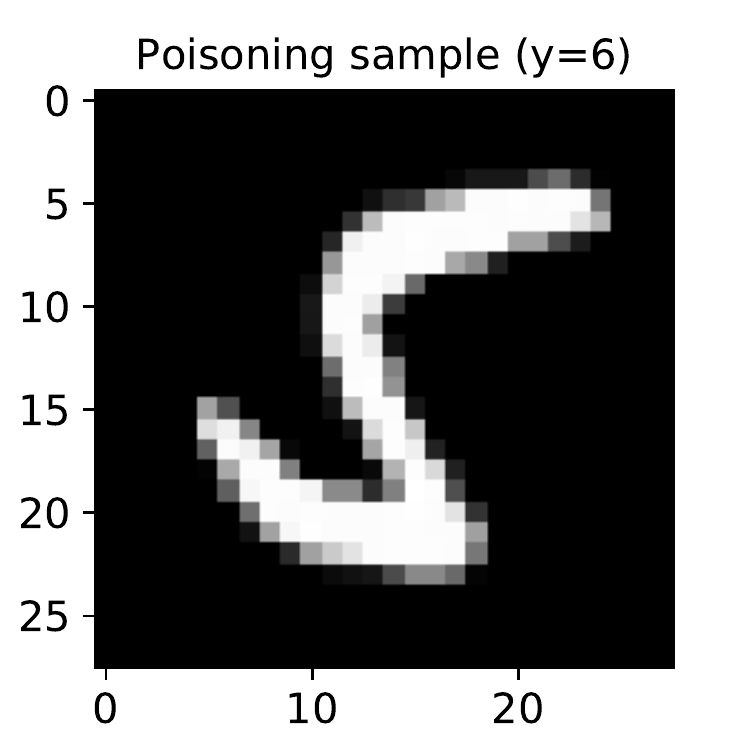}
	\includegraphics[width=0.15\textwidth]{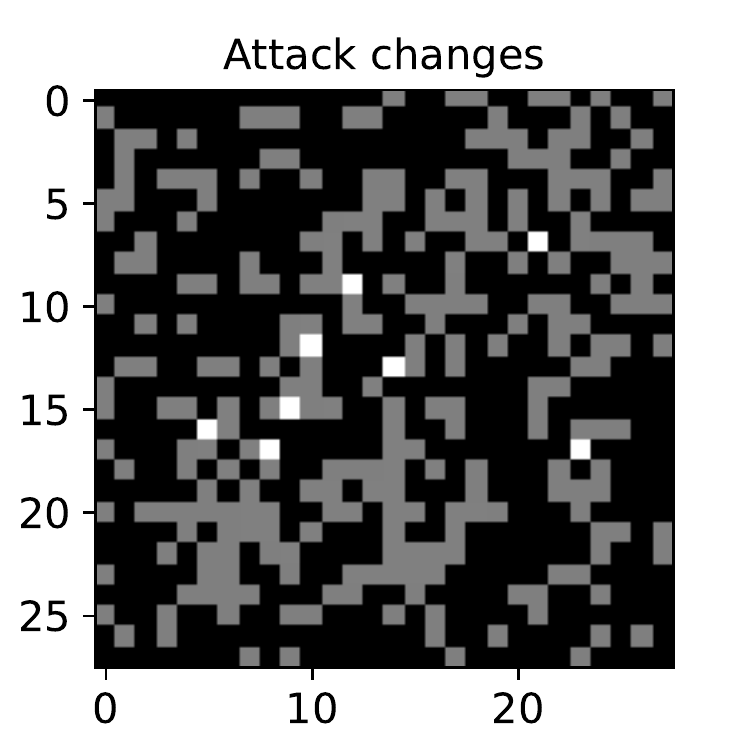}
\caption{Poisoning samples targeting the CNN.} \vspace{-5pt}
\label{fig:attack-digits-NN}
\end{figure}

\begin{figure}[t]
	\includegraphics[width=0.15\textwidth]{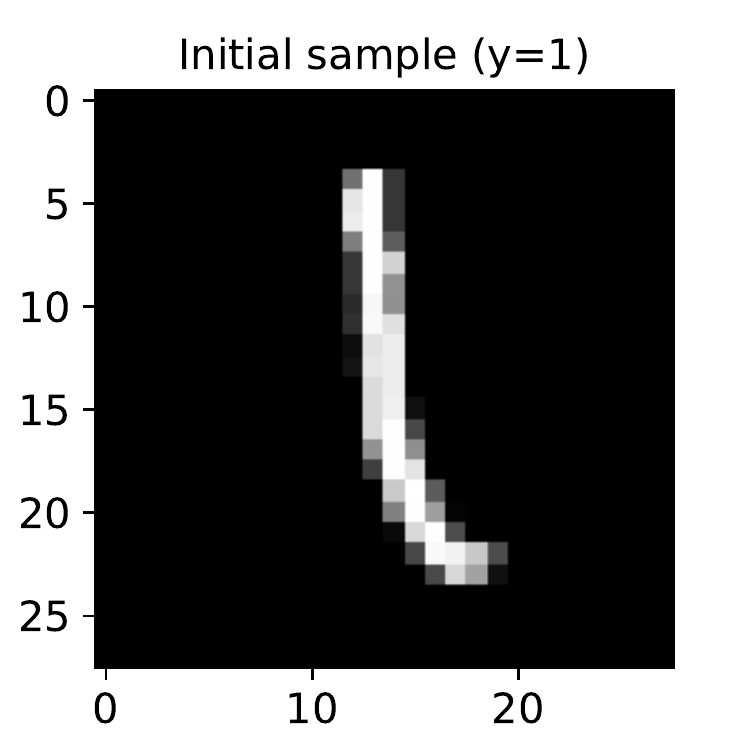}
	\includegraphics[width=0.15\textwidth]{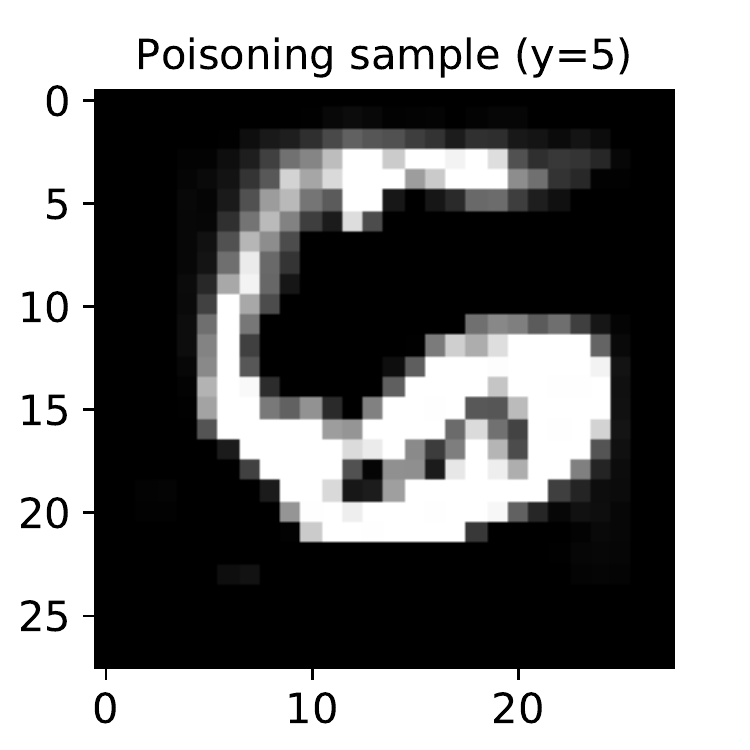}
	\includegraphics[width=0.15\textwidth]{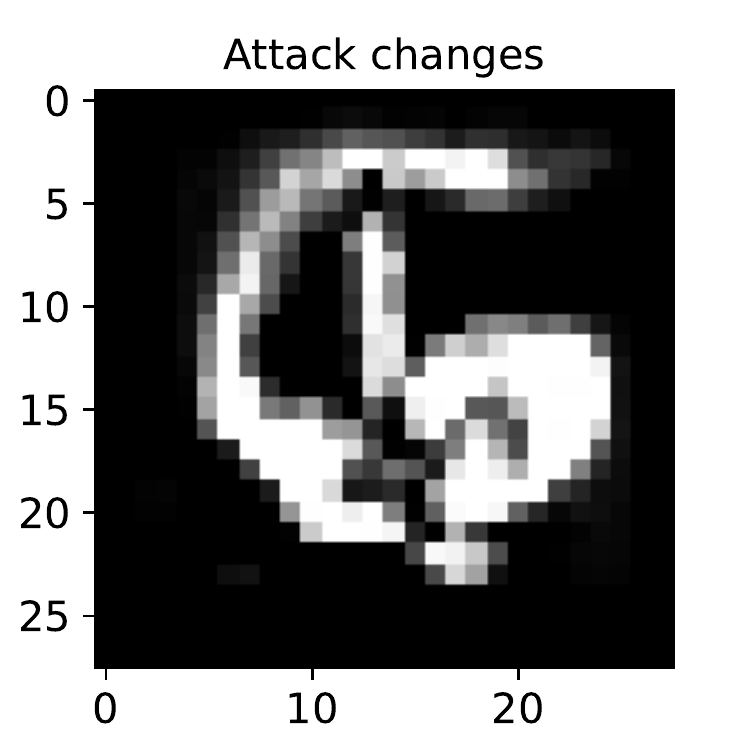}
	\includegraphics[width=0.15\textwidth]{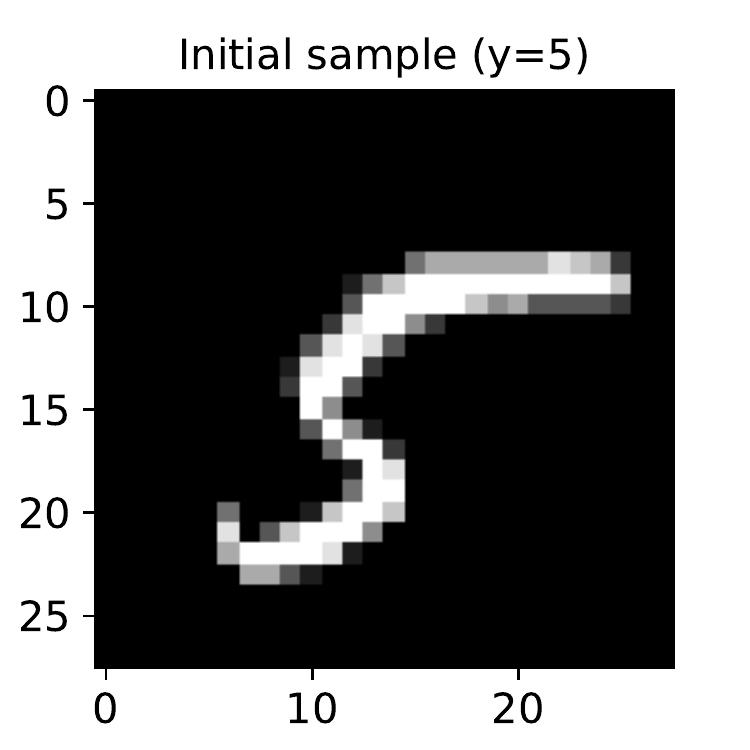}
	\includegraphics[width=0.15\textwidth]{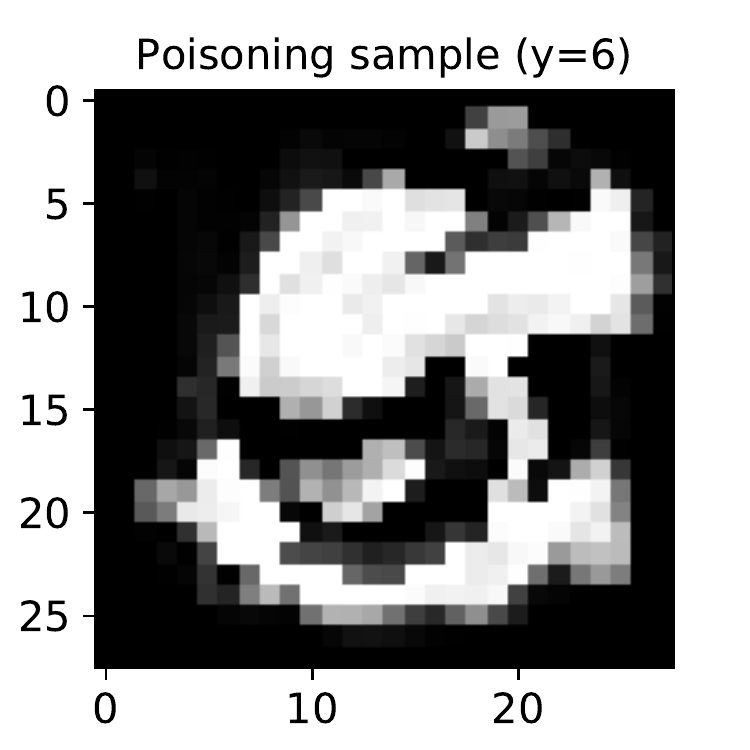}
	\includegraphics[width=0.15\textwidth]{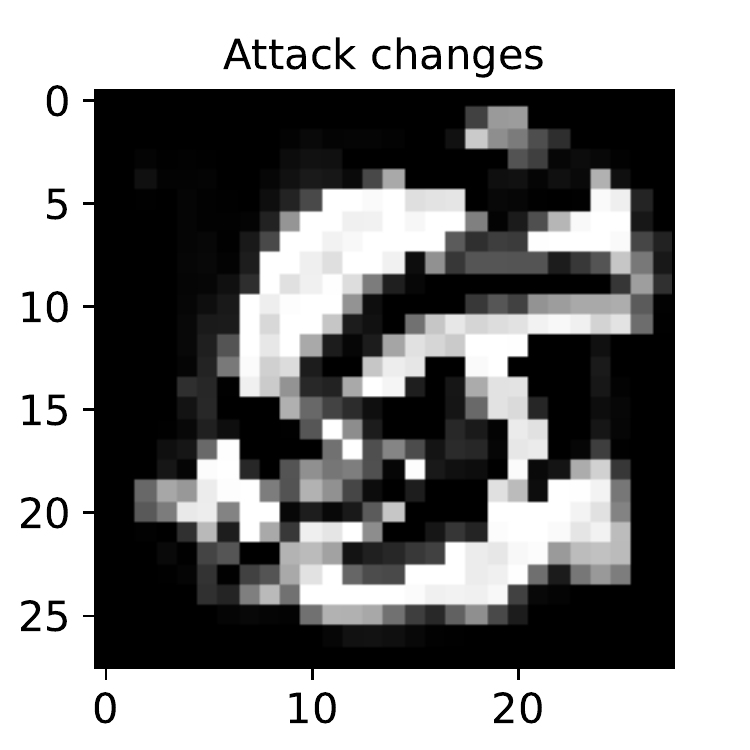}
	\includegraphics[width=0.15\textwidth]{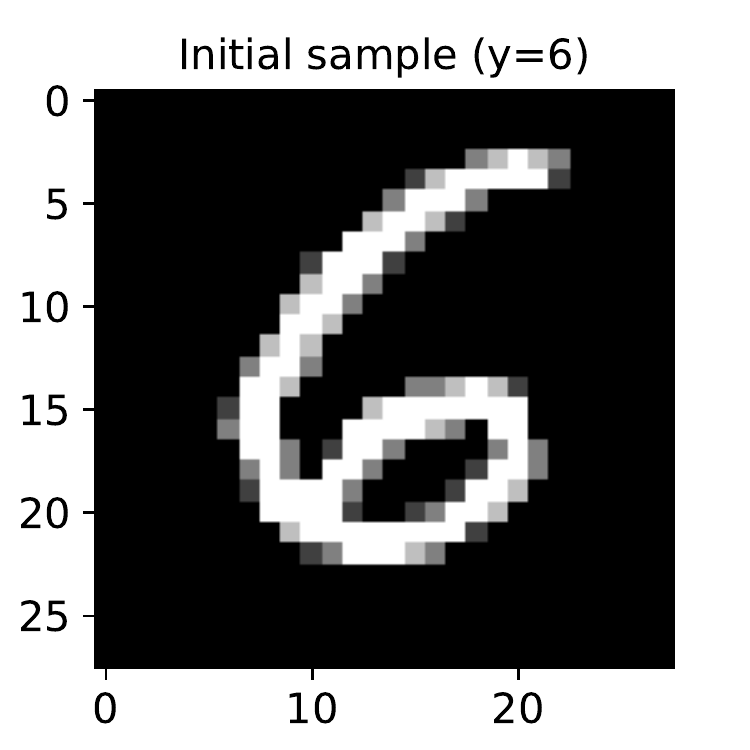}
	\includegraphics[width=0.15\textwidth]{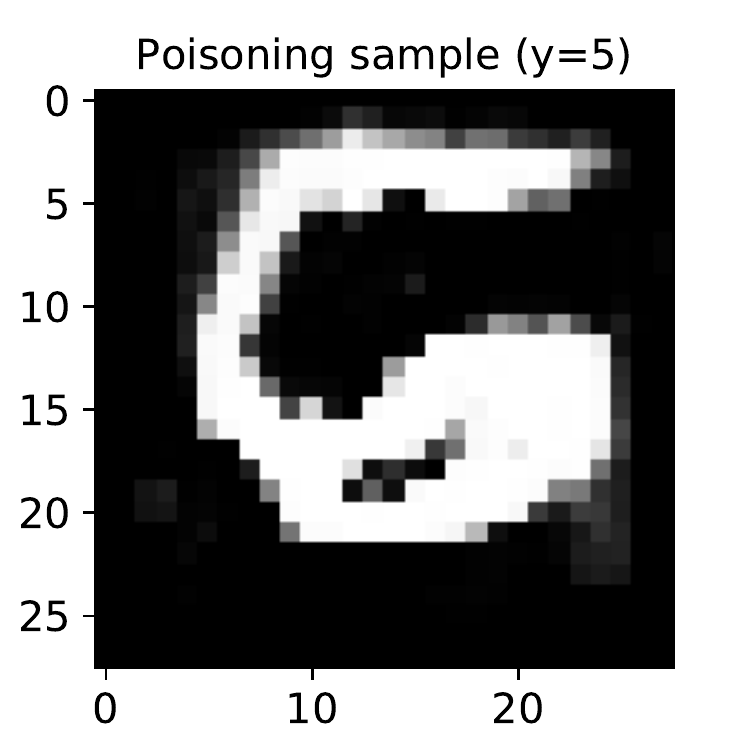}
	\includegraphics[width=0.15\textwidth]{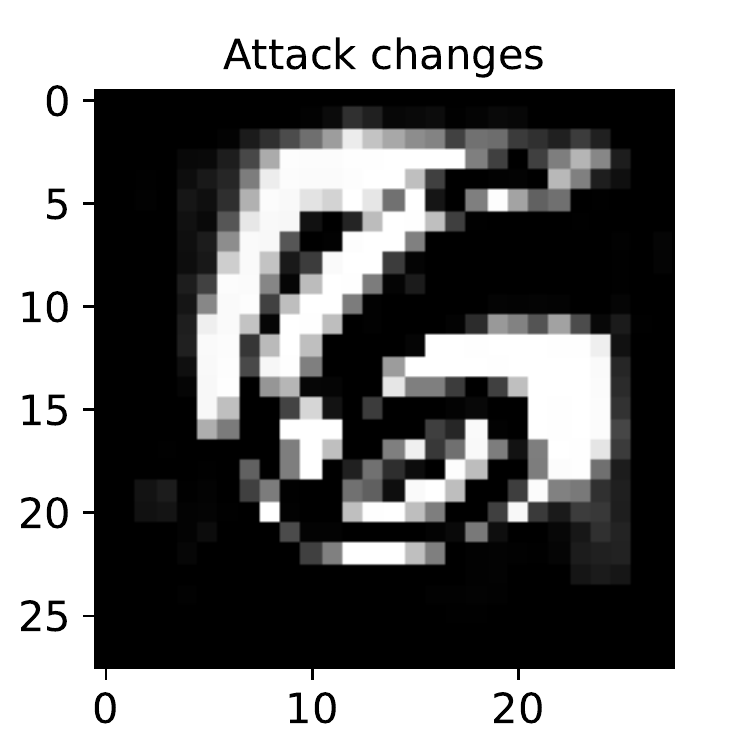}
\caption{Poisoning samples targeting the LR.} \vspace{-10pt}
\label{fig:attack-digits-LR}
\end{figure}

\myparagraph{Poisoning Deep Neural Networks} We finally report a proof-of-concept experiment to show the applicability of our attack algorithm to poison a deep network in an \emph{end-to-end} manner, \ie, accounting for all weight updates in each layer (instead of using a surrogate model trained on a frozen deep feature representation~\cite{koh17-icml}). 
To this end, we consider the convolutional neural network (CNN) proposed in~\cite{lecun98} for classification of the MNIST digit data, which requires optimizing more than $450,000$ parameters.\footnote{We use the implementation available at \url{https://github.com/tflearn/tflearn/blob/master/examples/images/convnet_mnist.py}.}
In this proof-of-concept attack, we inject $10$ poisoning points into the training data, and repeat the experiment on $5$ independent data splits, considering $1,000$ samples for training, and $2,000$ for validation and testing. For simplicity, we only consider the classes of digits $1$, $5$, and $6$ in this case.
We use Algorithm~\ref{alg:poisoning} to craft each single poisoning point, but, similarly to~\cite{biggio15-icml}, we optimize them iteratively, making $2$ passes over the whole set of poisoning samples. 
We also use the line search exploited in~\cite{biggio15-icml}, instead of a fixed gradient step size, to reduce the attack complexity (\ie, the number of training updates to the deep network).
Under this setting, however, we find that our attack points only slightly increase the classification error, though not significantly, while random label flips do not have any substantial effect.
For comparison, we also attack a multiclass LR classifier under the same setting, yielding an increase of the error rate from $2\%$ to $4.3\%$ with poisoning attacks, and to only $2.1\%$ with random label flips.
This shows that, at least in this simple case, deep networks seem to be more resilient against (a very small fraction of) poisoning attacks (\ie, less than $1\%$). Some of the poisoning samples crafted against the CNN and the LR are shown in Figs.~\ref{fig:attack-digits-NN} and~\ref{fig:attack-digits-LR}.
We report the initial digit (and its true label $y$), its poisoned version (and its label $y_{c}$), and the difference between the two images, in absolute value (rescaled to visually appreciate the modified pixels).
Notably, similarly to adversarial test examples, also poisoning samples against deep networks are visually indistinguishable from the initial image (as in~\cite{koh17-icml}), while this is not the case when targeting the LR classifier. This might be due to the specific shape of the decision function learned by the deep network in the input space, as explained in the case of adversarial test examples~\cite{szegedy14-iclr,goodfellow15-iclr}.
We however leave a more detailed investigation of this aspect to future work, along with a more systematic security evaluation of deep networks against poisoning attacks.
We conclude this section with a simple \emph{transferability} experiment, in which we use the poisoning samples crafted against the LR classifier to attack the CNN, and vice-versa. In the former case, the attack is totally ineffective, while in the latter case it has a similar effect to that of random label flips (as the minimal modifications to the CNN-poisoning digits are clearly irrelevant for the LR classifier). 


\section{Related Work}
\label{sec:RelatedWork}

Seminal work on the analysis of supervised learning in the presence of \emph{omniscient} attackers that can compromise the training data has been presented in \cite{kearns93,bshouty99}. While their results show the infeasibility of learning in such settings, their analysis reports an overly-pessimistic perspective on the problem. The first practical poisoning attacks against two-class classification algorithms have been proposed in \cite{nelson08,kloft12b}, in the context of spam filtering and anomaly detection. However, such attacks do not easily generalize to different learning algorithms.
More systematic attacks, based on the exploitation of KKT conditions to solve the bilevel problem corresponding to poisoning attacks have been subsequently proposed in~\cite{biggio12-icml,biggio15-icml,mei15-aaai,koh17-icml}. In particular, \citet{biggio12-icml} have been the first to demonstrate the vulnerability of SVMs to poisoning attacks. Following the same approach, \citet{biggio15-icml} have shown how to poison {LASSO}, {ridge regression}, and the {elastic net}. Finally, \citet{mei15-aaai} has systematized such attacks under a unified framework to poison convex learning algorithms with Tikhonov regularizers, based on the concept of machine teaching~\cite{zhu2013,patil}. The fact that these techniques require full re-training of the learning algorithm at each iteration (to fulfil the KKT conditions up to a sufficient finite precision), along with the intrinsic complexity required to compute the corresponding gradients, makes them too computationally demanding for several practical settings. Furthermore, this limits their applicability to a wider class of learning algorithms, including those based on gradient descent and subsequent variants, like deep neural networks, as their optimization is often truncated prior to meeting the stationarity conditions with the precision required to compute the poisoning gradients effectively. 
%
%
Note also that, despite recent work~\cite{koh17-icml} has provided a first proof of concept of the existence of \emph{adversarial training examples} against deep networks, this has been shown on a binary classification task using a surrogate model (attacked with standard KKT-based poisoning). In particular, the authors have  generated the poisoning samples by attacking a logistic classifier trained on the features extracted from the penultimate layer of the network (which have been kept fixed). 
Accordingly, to our knowledge, our work is thus the first to show how to poison a deep neural network in an \emph{end-to-end} manner, considering all its parameters and layers, and without using any surrogate model. 
%
%
Notably, our work is also the first to show (in a more systematic way) that poisoning samples can be \emph{transferred} across different learning algorithms, using \emph{substitute} (\aka \emph{surrogate}) models, as similarly demonstrated for evasion attacks (\ie, adversarial test examples) in~\cite{biggio13-ecml,srndic14} against SVMs and NNs, and subsequently in~\cite{papernot17-asiaccs} against deep networks.

\section{Conclusions, Limitations and Future Work}
\label{sec:Conclusion}

Advances in machine learning have led to a massive use of data-driven technologies with emerging applications in many different fields, including cybersecurity, self-driving cars, data analytics, biometrics and industrial control systems. At the same time, the variability and sophistication of cyberattacks have tremendously increased, making machine learning systems an appealing target for cybercriminals~\cite{barreno06-asiaccs,huang11}.

In this work, we have considered the threat of training data poisoning, \ie, an attack in which the training data is purposely manipulated to maximally degrade the classification performance of learning algorithms. While previous work has shown the effectiveness of such attacks against binary learners~\cite{biggio12-icml,mei15-aaai,biggio15-icml,koh17-icml}, in this work we have been the first to consider poisoning attacks in multiclass classification settings. To this end, we have extended the commonly-used threat model proposed in \cite{barreno06-asiaccs,barreno10,huang11} by introducing the concept of \emph{error specificity}, to denote whether the attacker aims to cause specific misclassification errors (\ie, misclassifying samples as a specific class), or generic ones (\ie, misclassifying samples as any class different than the correct one). 

Another important contribution of this work has been to overcome the limitations of state-of-the-art poisoning attacks, which require exploiting the stationarity (KKT) conditions of the attacked learning algorithms to optimize the poisoning samples~\cite{biggio12-icml,mei15-aaai,biggio15-icml,koh17-icml}. 
As discussed throughout this work, this requirement, as well as the intrinsic complexity of such attacks, limits their application only to a reduced class of learning algorithms. 
In this work, we have overcome these limitations by proposing a novel poisoning algorithm based on back-gradient optimization~\cite{domke12-aistats,maclaurin15-icml,pedregosa16-icml}.
Our approach can be applied to a wider class of learning algorithms, as it only requires the learning algorithm to update smoothly its parameters during training, without even necessarily fulfilling the optimality conditions with very high precision. Moreover, the gradients can be accurately estimated with the parameters obtained from an incomplete optimization of the learning algorithm truncated to a reduced number of iterations.
This enables the efficient application of our attack strategy to large neural networks and deep learning architectures, as well as any other learning algorithm trained through gradient-based procedures.
Our empirical evaluation on spam filtering, malware detection, and handwritten digit recognition has shown that neural networks can be significantly compromised even if the attacker only controls a small fraction of training points. 
We have also empirically shown that poisoning samples designed against one learning algorithm can be rather effective also in poisoning another algorithm, highlighting an interesting \emph{transferability} property, as that shown for evasion attacks (\aka adversarial test examples)~\cite{biggio13-ecml,srndic14,papernot17-asiaccs}. 

The main limitation of this work is that we have not run an extensive evaluation of poisoning attacks against deep networks, to thoroughly assess their security to poisoning. Although our preliminary experiments seem to show that they can be more resilient against this threat than other learning algorithms, a more complete and systematic analysis remains to be performed. Therefore, we plan to more systematically investigate the effectiveness of our back-gradient poisoning attack against deep networks in the very near future.
Besides the extension and evaluation of this poisoning attack strategy to different deep learning architectures and nonparametric models, further research avenues include: the investigation of the existence of \emph{universal perturbations} (not dependent on the initial attack point) for poisoning samples against deep networks, similarly to the case of universal adversarial test examples~\cite{goodfellow15-iclr,moosavi17-cvpr}; and the evaluation of defense mechanisms against poisoning attacks, through the exploitation of data sanitization and robust learning algorithms~\cite{rubinstein09,biggio11-mcs,steinhardt17-arxiv}.


\end{document}